%% file: main.tex
\documentclass[10pt,leqno]{amsart}
\usepackage[utf8]{inputenc}
\usepackage{graphicx}
\usepackage{indentfirst,csquotes}
\usepackage{pgfplots}
\usepackage{tcolorbox}
\usepackage{amssymb,amsthm,amsmath}
\usepackage{xcolor,paralist,hyperref,fancyhdr,etoolbox} 
\usepackage{braket}
\usepackage{subcaption} 
\usepackage{booktabs} 
\usepackage{lmodern}
\usepackage{import}
\usepackage{algorithm}
\usepackage{algpseudocode}
\usepackage{mathtools}
\usepackage[T1]{fontenc}
\usepackage{braket}
\usepackage{minted}
\usepackage{standalone}
\usepackage{tikz}
\usetikzlibrary{
  positioning,
  arrows.meta,
  shapes.geometric,
  decorations.pathreplacing,
  calc,
  shadows,        
  patterns,
  fit,
  backgrounds
}

\usepackage{xcolor}

\usepackage{amsmath}
\usepackage{tcolorbox}   
\usepackage{listings}    

\usetikzlibrary{positioning, arrows.meta, shapes.geometric}

\tikzstyle{block} = [rectangle, draw, text centered, rounded corners, minimum height=2em]
\tikzstyle{line} = [draw, -stealth, thick]
\tikzstyle{cloud} = [ellipse, draw, text centered, minimum height=2em, thick]

\usetikzlibrary{automata, positioning}
\usetikzlibrary{shapes.geometric, arrows, positioning}

\definecolor{roadcol}   {RGB}{80,80,80}
\definecolor{roadlight} {RGB}{160,160,160}
\definecolor{bridgecol} {RGB}{55,96,145}
\definecolor{bridgedark}{RGB}{35,65,105}
\definecolor{watercol}  {RGB}{100,170,210}
\definecolor{waterlight}{RGB}{170,215,235}
\definecolor{bankcol}   {RGB}{160,130,80}
\definecolor{banklight} {RGB}{200,175,120}
\definecolor{pillarcol} {RGB}{130,110,80}
\definecolor{cablecol}  {RGB}{180,50,50}
\definecolor{agentcol}  {RGB}{50,140,90}
\definecolor{agentbg}   {RGB}{220,245,225}
\definecolor{labelcol}  {RGB}{30,60,100}
\definecolor{budgetcol} {RGB}{200,100,30}
\definecolor{budgetbg}  {RGB}{255,240,220}
\definecolor{condgood}  {RGB}{60,150,80}
\definecolor{condfair}  {RGB}{210,160,30}
\definecolor{condpoor}  {RGB}{190,60,50}
\definecolor{cmdcol}    {RGB}{180,60,30}

\everymath=\expandafter{\the\everymath\displaystyle}

\definecolor{prismkw}{RGB}{0,0,180}
\definecolor{prismcmt}{RGB}{100,100,100}
\lstdefinelanguage{prism}{
  morekeywords={mdp, dtmc, module, endmodule, formula,
                label, const, init, rewards, endrewards,
                bool, int, double, true, false},
  sensitive=true,
  morecomment=[l]{//},
  morecomment=[s]{/*}{*/},
  keywordstyle=\color{prismkw}\bfseries,
  commentstyle=\color{prismcmt}\itshape,
}

\tikzstyle{block} = [rectangle, draw, text centered, rounded corners, minimum height=2em]
\tikzstyle{line} = [draw, -stealth, thick]
\tikzstyle{cloud} = [ellipse, draw, text centered, minimum height=2em, thick]
\tikzstyle{dashedcloud} = [ellipse, draw, dashed, text centered, minimum height=2em, thick]
\usetikzlibrary{positioning, arrows.meta, shapes.geometric, decorations.pathreplacing}

\tikzstyle{startstop} = [rectangle, rounded corners, minimum width=1.5cm, minimum height=0.5cm,text centered, draw=black, fill=red!30]
\tikzstyle{io} = [trapezium, trapezium left angle=70, trapezium right angle=110, minimum width=1cm, minimum height=0.5cm, text centered, draw=black, fill=blue!30]

\tikzstyle{process} = [rectangle, minimum width=3cm, minimum height=0.5cm, text centered, draw=black, fill=orange!30]
\tikzstyle{decision} = [diamond, minimum width=0.5cm, minimum height=0.1cm, text centered, draw=black, fill=green!30]
\tikzstyle{process2} = [rectangle, minimum width=1cm, minimum height=0.5cm, text centered, draw=black, fill=orange!30]
\tikzstyle{arrow} = [thick,->,>=stealth]
\usetikzlibrary{shapes,shadows,arrows,positioning,angles,quotes}
\tikzset{My Arrow Style/.style={single arrow, fill=black!15, anchor=base, align=center,text width=2.3cm}}
\tikzstyle{arrow} = [thick,->,>=stealth]
\usetikzlibrary{calc,trees,positioning,arrows,fit,shapes,calc}

\newtheorem{theorem}{Theorem}
\newtheorem{definition}[theorem]{Definition}

\hypersetup{
  colorlinks=true,
  linkcolor=blue,    
  citecolor=blue,    
  filecolor=blue,    
  urlcolor=blue      
}

\begin{document}

\title[Verifying and Explaining RL Policies for Multi-bridge Network Maintenance]{COOL-MC: Verifying and Explaining RL Policies for Multi-bridge Network Maintenance}
\author[Dennis Gross]{Dennis Gross}
\thanks{Email:\quad\texttt{dennis@artigo.ai}  \quad\textbar\quad \emph{COOL-MC}:\quad \url{https://github.com/LAVA-LAB/COOL-MC}}
\maketitle


\begin{abstract}
Aging bridge networks require proactive, verifiable, and interpretable maintenance strategies, yet \emph{reinforcement learning (RL)} policies trained solely on reward signals provide no formal safety guarantees and remain opaque to infrastructure managers.
We demonstrate \emph{COOL-MC} as a tool for verifying and explaining RL policies for multi-bridge network maintenance, building on a single-bridge \emph{Markov decision process (MDP)} from the literature and extending it to a parallel network of three heterogeneous bridges with a shared periodic budget constraint, encoded in the PRISM modeling language. We train an RL agent on this MDP and apply probabilistic model checking and explainability methods to the induced \emph{discrete-time Markov chain (DTMC)}, which arises from the interaction of the learned policy with the underlying MDP.
Probabilistic model checking reveals that the trained policy has a safety-violation probability of 3.5\% over the planning horizon, being slightly above the theoretical minimum of 0\% and indicating the suboptimality of the learned policy, noting that these results are based on artificially constructed transition probabilities and deterioration rates rather than real-world data, so absolute performance figures should be interpreted with caution.
The explainability analysis further reveals, for instance, a systematic bias in the trained policy toward the state of bridge 1 over the remaining bridges in the network. These results demonstrate \emph{COOL-MC}'s ability to provide formal, interpretable, and practical analysis of RL maintenance policies.
\end{abstract}


\section{Introduction}
Transportation infrastructure forms the backbone of modern society, enabling the movement 
of people, goods, and services that sustain economic growth and quality of life~\cite{bartle2017transportation}. 
Among the most critical yet vulnerable components of road and rail networks, bridges are subject to accelerating deterioration, deferred maintenance, and insufficient inspection regimes~\cite{grieco2024index, zhang2022causes, wardhana2003analysis}.
The consequences are severe and well-documented: collapses such as the I-35W bridge in Minneapolis (2007) and the Morandi Bridge in Genoa (2018) have claimed dozens of lives, while incidents such as the Carola Bridge in Dresden (2024) serve as stark reminders that failure can strike without warning~\cite{national2008collapse,domaneschi2020collapse,DW2024Dresden}.
All of these cases are symptoms of a global problem: across the world, a significant share of bridges are approaching or exceeding their design life, with repair backlogs running into the hundreds of billions of dollars~\cite{grieco2024index, fu2013statistical}.
Addressing this challenge requires systematic, proactive strategies for managing bridge networks throughout their lifecycles~\cite{yang2020life}.

In a world where resources are limited, it is necessary to adopt rational maintenance management approaches to support decision-makers in prioritising interventions across bridge portfolios~\cite{grieco2024index}.
One principled framework for this are \emph{Markov decision process models (MDPs)}, which are frequently employed in bridge maintenance~\cite{gopal1991application,robelin2007history,tao1994reliability,scherer1994markovian,wei2020optimal,hawk1995bridgit,MirzaeiZ,thompson1998pontis,morato2022optimal}: at each time step, 
a decision-maker observes the current condition of the bridge network, selects maintenance actions, and receives a reward reflecting the trade-off between intervention costs and structural survival, while each bridge transitions to a new condition state according to probabilistic deterioration dynamics~\cite{baier2008principles}.
Unfortunately, solving MDPs optimally via \emph{probabilistic model checking}, which verifies policy guarantees by exhaustively analyzing the full MDP~\cite{baier2008principles}, becomes computationally intractable as the number of structures grows, since the joint state and action space expands exponentially, a challenge known as the \emph{curse of dimensionality}~\cite{lu2024overcoming}.
\emph{Reinforcement learning (RL)} addresses this challenge by training an agent to learn a maintenance policy through repeated interactions with a simulated deterioration MDP~\cite{schulman2017proximal, DBLP:journals/corr/MnihKSGAWR13}: the agent observes the bridge network state, decides which bridges to repair, or replace within the available budget, and receives a reward signal that incentivizes long-term structural performance.
Modern RL algorithms represent this policy using a \emph{neural network (NN)}, enabling the agent to generalize across large MDPs~\cite{DBLP:journals/corr/MnihKSGAWR13}.

\begin{figure}[t]
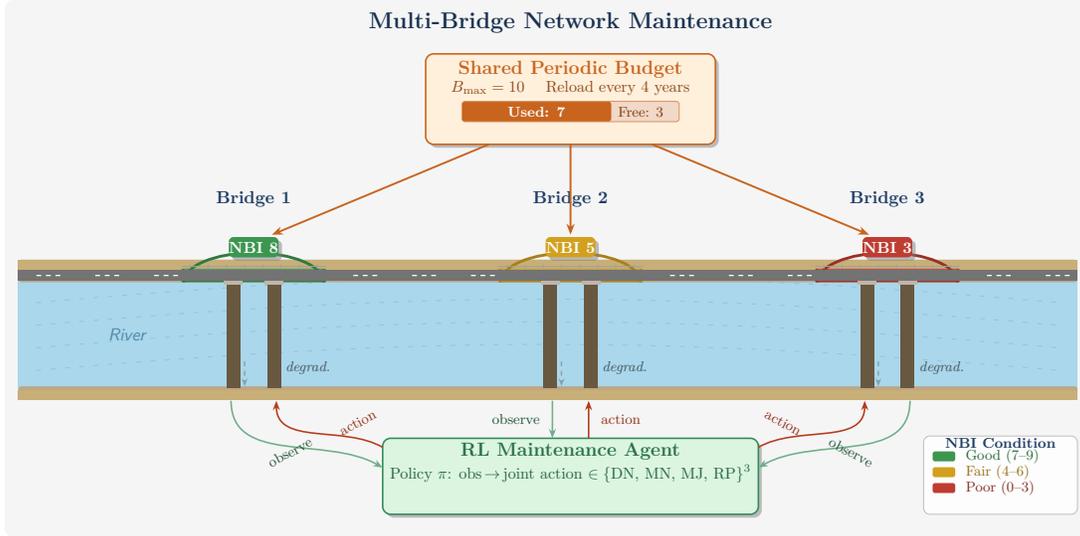

  \centering
  \includestandalone[width=\textwidth]{bridge_network}
  \caption{%
    The multi-bridge network maintenance problem.
    Three heterogeneous bridges span a shared river, each rated
    on the NBI condition scale (green = Good, amber = Fair, red = Poor).
    A shared periodic budget ($B_{\max}=10$, reloaded every four years)
    is allocated by the RL maintenance agent, which observes all bridge
    conditions and issues joint maintenance actions
    $\pi \in \{\text{Do Nothing (DN)},\text{Minor Maintenance (MN)},\text{Major Maintenance (MJ)},\\ \text{Replacement (RP)}\}^3$ at every time step.
  }
  \label{fig:bridge-network}
\end{figure}

However, RL policies can exhibit \emph{unsafe behavior}~\cite{DBLP:conf/setta/GrossJJP22} 
such as replacing fully functional bridges, as reward signals are often insufficient to capture complex structural 
safety requirements~\cite{wei2020optimal}.
NN-based policies are additionally \emph{hard to interpret}, as the internal complexity of NNs hides the reasoning behind each maintenance decision~\cite{DBLP:journals/ml/Bekkemoen24}.
In the bridge maintenance setting, this opacity is particularly problematic: infrastructure managers need to understand \emph{why} a policy recommends a given action for a specific bridge and whether it appropriately accounts for the broader network's condition.
Without such understanding, adoption of RL-based maintenance policies in practice remains challenging~\cite{vouros2022explainable,DBLP:journals/ml/Bekkemoen24}.

The tool \emph{COOL-MC} addresses both of these challenges by combining RL with probabilistic model checking and explainability~\cite{DBLP:conf/setta/GrossJJP22,gross2025pctl,DBLP:conf/esann/GrossS24,gross2024enhancing}: After learning a policy via RL, it constructs only the reachable state space induced by that policy, resolving all nondeterminism and yielding a \emph{discrete-time Markov 
chain (DTMC)} amenable to probabilistic model checking and explainable RL methods~\cite{DBLP:conf/setta/GrossJJP22}.
By constructing only the reachable states under the trained policy rather than enumerating the full state space, this approach mitigates the curse of dimensionality 
that renders classical methods such as full-MDP model checking intractable~\cite{DBLP:conf/setta/GrossJJP22}.

To date, however, \emph{COOL-MC}'s capabilities~\cite{DBLP:conf/setta/GrossJJP22,gross2026formallyverifyingexplainingsepsis,gross2026coolmcverifyingexplainingrl} have not been applied to the verification and explanation of RL policies for multi-bridge network maintenance.

In this paper, we encode a network of three heterogeneous bridges as an MDP, extending the single-bridge formulation of Wei et al.~\cite{wei2019reinforcement} to a parallel network with a shared periodic budget constraint and the full ten-point NBI condition scale.
We train an RL agent on this MDP and apply \emph{COOL-MC} to the induced DTMC $D^\pi$ to carry out a policy analysis spanning probabilistic model checking~\cite{DBLP:conf/setta/GrossJJP22}, feature-level explainability~\cite{gross2025pctl}, action-level behavioral profiling~\cite{gross2025pctl}, and counterfactual what-if analysis~\cite{gross2024enhancing}.
Probabilistic model checking reveals, for instance, that the trained policy has a safety-violation probability of 3.5\% over the 20-year planning horizon, being slightly above the theoretical minimum of 0\%, with the caveat that these results are based on artificially constructed transition probabilities and deterioration rates rather than real-world data.
The explainability analysis further uncovers, for instance, that the trained policy is biased toward the state of bridge~1, attributing it significantly more importance than the other bridges in the network.

\textbf{Our main contribution} is the demonstration of \emph{COOL-MC} as a tool for multi-bridge network maintenance, combining RL with probabilistic model checking and explainability to provide a formal analysis of a bridge network maintenance RL policy, including verified safety and performance properties, feature-level explanations, action-level behavioral profiling, and counterfactual what-if analysis of the trained RL maintenance policy, thereby showcasing \emph{COOL-MC}'s capabilities for maintenance tasks.

\section{Related Work}
Maintaining a bridge network in a safe condition requires, as a first step, reliable knowledge of the condition of each structure.
Bridge condition assessment in the USA is governed by the National Bridge Inventory,  which requires inspectors to rate each bridge on a 0-9 integer scale, where a rating of 7 or above classifies the structure as being in good condition, ratings of 5 or 6 as fair, and a rating of 4 or below on any primary component as poor~\cite{fhwa1995recording}. Different rating systems exist across countries~\cite{matos2023comparison}, but in our work, we adhere to this convention.

Bridge condition assessment relies on a range of inspection methods, from routine visual inspections and hands-on tactile examination to advanced non-destructive evaluation techniques such as ground-penetrating radar, acoustic emission monitoring, and unmanned aerial vehicle surveys~\cite{graybeal2002visual,nair2010acoustic,zhang2022towards, ahmed2020review}.
Empirical studies have nonetheless demonstrated that even trained inspectors exhibit considerable inter-inspector variability, with condition ratings for the same component differing by multiple scale points across inspectors~\cite{graybeal2002visual}, and subsurface deterioration processes such as internal corrosion and scour remain difficult to detect through any routine inspection method~\cite{fhwa2022nbis,madanat1994optimal}.
These uncertainties would, in principle, motivate a \emph{partial observable MDP (POMDP)} formulation~\cite{ellis1995inspection,papakonstantinou2014optimum} in which the agent maintains a belief distribution over the true structural state~\cite{andriotis2019managing}.
However, a POMDP formulation requires explicit observation matrices and component-level transition models under partial observability, which in turn demands precision estimates of inspections across the country and long-term monitoring records that are rarely available for real bridge networks~\cite{andriotis2019managing}. Under the MDP setting, by contrast, no model for component observations is required at all~\cite{andriotis2019managing}, which makes full observability a practically grounded modelling choice when such data are absent.
We therefore follow the convention of prior RL-based bridge maintenance studies~\cite{wei2020optimal,andriotis2019managing} and assume that condition ratings faithfully reflect the true state of each bridge.
Under this assumption, COOL-MC can construct the full reachable state space induced by the trained policy and provide both formal verification guarantees and rigorous behavioural explanations for this modelled MDP.

To obtain such policies, dynamic and linear programming algorithms are frequently employed~\cite{kuhn2010network,medury2014simultaneous,robelin2006dynamic}.
However, these algorithms are expensive and inefficient for problems with large state or action spaces~\cite{wei2020optimal}.
It is precisely this scalability barrier that motivated the adoption of RL for infrastructure maintenance~\cite{arcieri2024pomdp,li2024research,kazemeini2023identifying,yao2022large,han2021asphalt,yao2020deep,renard2021minimizing,asghari2025network,liu2025multi,thangeda2024infralib,jing2026intelligent,van2025deep,cheng2024knowledge,lai2024synergetic,taherkhani2024towards,hamida2023hierarchical,morato2023inference,cheng2021decision,andriotis2019managing,wei2020optimal,yang2022adaptive,yang2022deep,lei2022deep,du2022parameterized,zhou2022advanced,zhou2022reinforcement,dong2022deep,lei2023sustainable,saifullah2024multi,latifi2021deep,barua2022planning,mohammadi2022deep,sresakoolchai2023railway,lee2024stochastic,botteghi2021towards,jimenez2024maintenance}.
RL agents learn maintenance policies directly from simulated interaction with a deterioration environment, approximating optimal policies with neural networks that generalize across the vast joint state spaces~\cite{andriotis2019managing,wei2020optimal}.

Unfortunately, RL training solely on rewards does not guarantee that safety requirements are automatically satisfied~\cite{wei2020optimal,DBLP:conf/setta/GrossJJP22}.
While formal verification methods have been applied to RL policies in different domains~\cite{ganai2023iterative,fisac2019bridging,fisac2018general,unniyankal2023rmlgym,marzari2025verifying,zolfagharian2024smarla,mannucci2023runtime,lazarus2020runtime,mallozzi2019runtime,jin2022trainify,akintunde2022formal,bacci2020probabilistic,dong2022dependability,bacci2022verified,bacci2021verifying,jin2021learning,tian2023boosting,corsi2021formal,tran2019safety,li2017reinforcement,aksaray2016q,le2024reinforcement,alur2023policy,cai2021reinforcement,hahn2020reward,bozkurt2020control,hasanbeig2019reinforcement,littman2017environment,sadigh2014learning,fu2014probably,wang2024safe,hunt2021verifiably,anderson2020neurosymbolic,junges2016safety,gross2025turn,shao2023sample,jansen2018machine,landers2023deep,DBLP:conf/sigcomm/EliyahuKKS21,DBLP:conf/sigcomm/KazakBKS19,DBLP:journals/corr/DragerFK0U15,DBLP:conf/pldi/ZhuXMJ19,DBLP:conf/seke/JinWZ22,bensalem2024bridging,schmittlearning,venkataraman2020tractable,gangopadhyay2021counterexample,wu2024verified,gross2026coolmcverifyingexplainingrl}, none have been directly applied in combination with explainable RL to infrastructure maintenance policies. This gap is consequential: without formal guarantees and interpretable explanations, infrastructure managers can neither verify that a policy meets safety requirements across all network states based on their model nor understand the reasoning behind individual maintenance decisions, both of which are prerequisites for practical adoption~\cite{vouros2022explainable,DBLP:journals/ml/Bekkemoen24}.
We address this by applying COOL-MC, which combines probabilistic model checking with explainable RL to provide both formal verification of policy objectives and transparent interpretation of policy behaviour~\cite{DBLP:conf/setta/GrossJJP22,gross2025pctl,DBLP:conf/esann/GrossS24,gross2024enhancing,gross2026formallyverifyingexplainingsepsis}.
To the best of our knowledge, this combination has not 
been explored in the infrastructure maintenance literature.

\section{Background}
First, we introduce \emph{Markov decision processes (MDPs)} as the formal framework for sequential decision-making. We then describe probabilistic model checking for verifying policy properties, RL, and different explainability methods for explaining their RL policy decisions.

\subsection{Probabilistic Systems}
A \textit{probability distribution} over a set $X$ is a function $\mu \colon X \rightarrow [0,1]$ with $\sum_{x \in X} \mu(x) = 1$. The set of all distributions on $X$ is $Distr(X)$.

\begin{definition}[MDP]\label{def:mdp}
A \emph{Markov decision process (MDP)} is a tuple $M = (S,s_0,Act,Tr, rew,\\AP,L)$
where $S$ is a finite, nonempty set of states; $s_0 \in S$ is an initial state; $Act$ is a finite set of actions; $Tr\colon S \times Act \rightarrow Distr(S)$ is a partial probability transition function and $Tr(s,a,s')$ denotes the probability of transitioning from state $s$ to state $s'$ when action $a$ is taken;
$rew \colon S \times Act \rightarrow \mathbb{R}$ is a reward~function;
$AP$ is a set of atomic propositions;
$L \colon  S \rightarrow 2^{AP}$ is a labeling~function.
\end{definition}
An \emph{agent} interacts with the MDP by observing the current state and selecting actions to maximize cumulative reward. The agent's behavior is governed by a \emph{policy}.

We represent each state $s \in S$ as a vector of $d$ features $(f_1, \dots, f_d)$, where $f_i \in \mathbb{Q}$.
The available actions in $s \in S$ are $Act(s) = \{a \in Act \mid Tr(s,a) \neq \bot\}$ where $Tr(s, a) = \bot$ means that action $a$ is not available in state $s$.
In our setting, we assume that all actions are available at all states.

\begin{definition}[DTMC]\label{def:dtmc}
A \emph{discrete-time Markov chain (DTMC)} is a tuple $D = (S, s_0, Tr, AP, L)$
where $S$, $s_0$, $AP$, and $L$ are as in Definition~\ref{def:mdp}, 
and $Tr \colon S \rightarrow Distr(S)$ is a probability transition~function.
\end{definition}
We write $Tr(s, s')$ to denote the probability of transitioning from state  $s$ to state $s'$ under $Tr$.

In many practical settings, the agent does not have direct access to the underlying state $s \in S$ of the MDP. Instead, the agent receives an observation that may represent a transformed view of the true state. We formalize this through an observation function.

\begin{definition}[Observation]
    We define the observation function $\mathbb{O} \colon S \rightarrow O$ as a function that maps a state $s \in S$ to an observation $o \in O$.
    An observation $o \in O$ is a vector of features $(f_1, \dots, f_d)$ where $f_j \in \mathbb{Q}$.
    The observed features may differ from the exact state features.
\end{definition}

A policy operates on observations rather than on the underlying states directly.
In our setting, the environment is fully observable, so the observation function $\mathbb{O}$ maps each state feature directly to its corresponding observation feature, i.e., $\mathbb{O}$ is the identity on the state~features.

\begin{definition}
    A \emph{memoryless deterministic policy $\pi$} for an MDP $M$ is a function $\pi \colon O \rightarrow Act$ that maps an observation $o \in O$ to action $a \in Act$.
\end{definition}

\begin{figure}[]
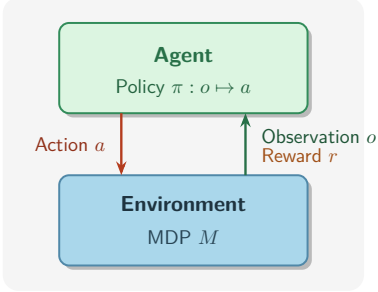

\centering
\scalebox{0.35}{
    \includestandalone[width=\textwidth]{rl}
    }
\caption{Sequential decision-making loop. The agent receives an observation and reward from the environment, selects an action according to its policy, and the environment transitions to a new state.}
\label{fig:rl}
\end{figure}

Applying a policy $\pi$ to an MDP $M$ with observation function $\mathbb{O}$ yields an \emph{induced DTMC} $D^\pi$ where all non-determinism is resolved: for each state $s$, the transition function becomes $Tr(s, s') = Tr(s, \pi(\mathbb{O}(s)), s')$.
The induced DTMC fully characterizes the observable behavior of the policy: the states visited, the transitions taken, and the probabilities of all outcomes.
The interaction between the policy and environment is depicted in Figure~\ref{fig:rl}.

\subsection{Probabilistic Model Checking}

\emph{Probabilistic model checking} enables the verification of quantitative properties of stochastic systems.
\emph{COOL-MC} uses \emph{Storm}~\cite{DBLP:journals/sttt/HenselJKQV22} internally as its model checker, which can verify properties of both MDPs and DTMCs.
Among the most fundamental properties are \emph{reachability} queries, which assess the probability of a system reaching a particular state.
For example, one might say: ``The reachability probability of reaching an unsafe state is~0.1.''

\begin{figure}[htbp]
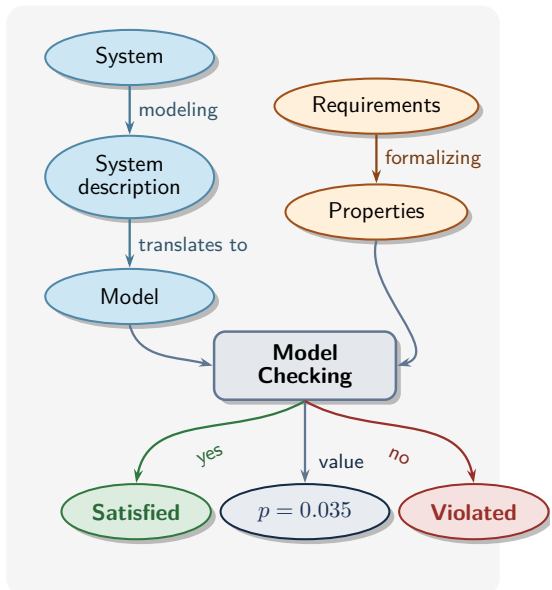

    \centering
    \scalebox{0.5}{
    \includestandalone[width=\textwidth]{model_checking}
    }
    \caption{General model checking workflow~\cite{DBLP:journals/sttt/HenselJKQV22}. The system is formally modeled, the requirements are formalized, and both are input to a model checker such as \emph{Storm}, which verifies the property.}
    \label{fig:model_checking}
\end{figure}

The general workflow for model checking is as follows (see also Figure~\ref{fig:model_checking}).
First, the system is formally modeled using a language such as \emph{PRISM}~\cite{prism_manual}.
Next, the property of interest is formalized in a temporal logic.
Using these inputs, the model checker verifies whether the property holds or computes the relevant probability.

The \emph{PRISM} language~\cite{prism_manual} describes probabilistic systems as a collection of \emph{modules}, each containing typed \emph{variables} and guarded \emph{commands}.
A state of the system is a valuation of all variables, and the state space $S$ is the set of all possible valuations.
Each command takes the~form
\[
\texttt{[action]}\ g \rightarrow \lambda_1 : u_1 + \cdots + \lambda_n : u_n
\]
where $g$ is a Boolean guard over the variables, each $\lambda_j$ is a probability, and each $u_j$ is an update that assigns new values to the variables.
When the system is in a state satisfying $g$, the command can be executed: update $u_j$ is applied with probability $\lambda_j$, transitioning the system to a new state.
For an MDP, each enabled command in a state corresponds to a distinct probabilistic choice, and the nondeterministic selection among enabled commands is resolved by a policy.
Listing~\ref{lst:prism_example} shows a simplified MDP in \emph{PRISM}.
The model declares a single module with a state variable \texttt{s} ranging over integer identifiers.
Each command specifies a transition: a guard on the current state, an action label, and a probabilistic update.
For instance, from state $s=0$, taking action \texttt{a0} leads to state $s=1$ with probability $0.7$ and to state $s=2$ with probability $0.3$.
States $s=2$ and $s=3$ are absorbing terminal states corresponding to different labels.
A reward structure assigns a reward of $+1$ to the $s=2$ state.
Additionally, \emph{PRISM} supports \emph{atomic labels} that annotate states with propositions.
These labels are defined using \texttt{label} declarations, which assign a name to a Boolean condition over the state variables.
For example, \texttt{label "goal" = s=2;} marks all states where $s=2$ with the proposition \textit{goal}.
Such labels serve as the atomic propositions referenced in PCTL properties.
\emph{COOL-MC} allows automated user-specified state labeling of \emph{PRISM} models.
For a complete formal treatment of the \emph{PRISM} semantics, we refer to~\cite{prism_manual}.

\begin{figure}[htbp]
\begin{lstlisting}[
    language={},
    basicstyle=\ttfamily\small,
    keywordstyle=\bfseries,
    commentstyle=\itshape\color{gray},
    frame=single,
    numbers=left,
    numberstyle=\tiny,
    caption={Simplified PRISM model. States are encoded as integers; transitions are guarded by the current state and labeled with actions. Atomic labels annotate states with propositions used in PCTL~properties.},
    label={lst:prism_example},
    escapeinside={(*}{*)}
]
mdp

module example
  s : [0..3] init 0; // state variable

  // From state 0, action a0:
  //   -> state 1 (w.p. 0.7), state 2 (w.p. 0.3)
  [a0] s=0 -> 0.7:(s'=1) + 0.3:(s'=2);

  // From state 0, action a1:
  //   -> state 1 (w.p. 0.4), state 3 (w.p. 0.6)
  [a1] s=0 -> 0.4:(s'=1) + 0.6:(s'=3);

  // From state 1, action a0:
  //   -> state 2 (w.p. 0.8), state 3 (w.p. 0.2)
  [a0] s=1 -> 0.8:(s'=2) + 0.2:(s'=3);

  // Absorbing states (survival and death)
  [end] s=2 -> 1.0:(s'=2); // goal
  [end] s=3 -> 1.0:(s'=3); // empty

endmodule

// Atomic labels for state propositions
label "goal" = s=2;
label "empty" = s=3;

// Reward: +1 for reaching goal state
rewards
  s=2 : 1;
endrewards
\end{lstlisting}
\end{figure}

Properties are specified using PCTL~\cite{hansson1994logic}, a branching-time temporal logic for reasoning about probabilities over paths.
In this paper, we use two path operators.
The \emph{eventually} operator $\lozenge\,\varphi$ (also written $F\,\varphi$) states that $\varphi$ holds at some future state along a path.
The \emph{until} operator $\varphi_1 \;\mathcal{U}\; \varphi_2$ states that $\varphi_1$ holds at every state along a path until a state is reached where $\varphi_2$ holds.
A bounded variant $\lozenge^{\leq B}\,\varphi$ restricts the eventually operator to paths of at most $B$ steps.
A PCTL property has the form
\[
P_{\sim p} (\psi)
\]
where $\psi$ is a path formula (such as $\lozenge\,\varphi$ or $\varphi_1 \;\mathcal{U}\; \varphi_2$), $\sim$ is a comparison operator ($<$, $\leq$, $\geq$, $>$), and $p \in [0,1]$ is a probability threshold.
Beyond checking whether a property is satisfied, \emph{Storm} can compute the exact probability, denoted $P_{=?}(\psi)$.
For example, $P_{=?}(\lozenge^{\leq 200}\;\textit{empty})$ computes the exact probability of reaching a stockout within 200~steps.

\subsection{Reinforcement Learning}
The standard goal of RL is to learn a policy $\pi$ for an MDP that maximizes the expected accumulated discounted reward~\cite{DBLP:journals/ml/Bekkemoen24}
\[
\mathbb{E}\left[\sum_{t=0}^{N} \gamma^t \cdot rew(s_t, a_t)\right],
\]
where $\gamma \in [0,1]$ is the discount factor, $rew(s_t, a_t)$ is the reward at time step $t$, and $N$ is the episode length.

\subsection{Explainability Methods}\label{sec:xrl}
Explainability methods aim to make trained RL policies understandable~\cite{DBLP:journals/csur/MilaniTVF24}.
Global methods interpret the overall policy's behavior~\cite{DBLP:conf/aiide/SieusahaiG21}, while local methods explain a policy's decision-making in individual states.
In this work, we use four complementary methods: \emph{feature lumping}, \emph{gradient-based saliency ranking}, \emph{action labeling}, and \emph{counterfactual action replacement}, each combined with PCTL model checking on the induced~DTMC.

Feature lumping~\cite{DBLP:conf/setta/GrossJJP22} provides a global sensitivity analysis by coarsening the observation space.
For a selected feature $f_i$, exact values are replaced by binned representatives so that the memoryless deterministic policy $\pi$ operates on a reduced observation $o \in O$.
The induced DTMC is reconstructed under the lumped policy on the original MDP $M$, and the PCTL query of interest is re-computed.
A negligible change in the verified probability indicates that $\pi$ is robust to coarsened information about $f_i$; a large change reveals that precise resolution of that feature is safety-critical.

Gradient-based saliency~\cite{samadi2024safe} complements lumping with a local, per-state measure of feature relevance.
For each state $s$ with observation $o = \mathbb{O}(s)$, the saliency of feature $f_i$ is $\lvert \partial \pi(o) / \partial f_i \rvert$, the absolute gradient of the policy output with respect to the $i$-th input.
Aggregating these scores over all visited states yields a global feature-importance ranking, and conditional rankings can be computed over restricted subsets of $D^\pi$.

Action labeling annotates each state in the induced DTMC with the joint maintenance action selected by the trained policy~\cite{gross2026formallyverifyingexplainingsepsis}.
These labels serve as atomic propositions that are referenceable in PCTL, enabling formal queries directly over the policy's maintenance behavior.

Counterfactual action replacement modifies the policy by substituting one action with another across all states where it is selected, then reconstructs and re-verifies the induced DTMC~\cite{gross2024enhancing}.
This enables targeted what-if analysis that isolates the effect of specific maintenance choices on safety outcomes without retraining the policy.

\section{Methodology}
Our methodology proceeds in four stages.
First, we encode the multi-bridge network maintenance problem as an MDP in the 
\emph{PRISM} modeling language, defining states, actions, transition dynamics, 
and reward structure that capture the joint maintenance planning problem 
(\S\ref{sec:prism-encoding}).
Second, we train a deep RL policy using PPO to learn a maintenance strategy that 
maximizes structural survival while minimizing intervention costs 
(\S\ref{subsec:rl-training}).
Third, we verify the trained policy on its induced DTMC by applying PCTL queries 
to assess safety properties such as bridge failure probability and budget utilization 
(\S\ref{subsec:policy-verification}).
Fourth, we explain the policy's decision-making through feature lumping, 
gradient-based saliency ranking, action labeling, and counterfactual action 
replacement, combined with PCTL queries on the induced DTMC 
(\S\ref{subsec:policy-explanation}).

\subsection{PRISM MDP Encoding}
\label{sec:prism-encoding}
We encode the multi-bridge network maintenance problem as an MDP
$M$ in the PRISM language.
The formulation is inspired by the single-component bridge MDP
of Wei et al.~\cite{wei2019reinforcement} with minor differences
(\S\ref{sec:differences}), adapting their deterioration model to
a parallel network of three bridges with a shared periodic budget
constraint and extending the condition representation to the full
ten-point NBI scale.

\subsubsection{State space}
Each state $s \in S$ is a feature vector $(f_1, \dots, f_7)$
with components
\begin{equation*}
    s = (\mathit{cond\_b1},\; \mathit{cond\_b2},\; \mathit{cond\_b3},\;
         \mathit{budget},\; \mathit{cycle\_year},\;
         \mathit{year},\; \mathit{init\_done}),
\end{equation*}
where each $\mathit{cond\_bi} \in \{0, 1, \ldots, 9\}$ is the
structural condition of bridge $i \in \{1,2,3\}$ on the full
National Bridge Inventory (NBI) scale~\cite{fhwa1995recording},
defined in Table~\ref{tab:condition-states}.
The NBI assigns integer ratings 0--9 to bridge elements based on
visual inspection, where 9 denotes Excellent condition and 0
denotes a Failed structure.
The PRISM variable directly equals the NBI rating:
$\mathit{cond\_bi} = \text{NBI rating}$,
so that higher values represent better condition, matching the
natural NBI ordering.
$\mathit{budget} \in \{0,\ldots,B_{\max}\}$ is the remaining
maintenance budget within the current four-year cycle;
$\mathit{cycle\_year} \in \{0,1,2,3\}$ is the year-within-cycle
counter;
$\mathit{year} \in \{0,\ldots,T_{\max}\}$ is the global planning
year; and $\mathit{init\_done} \in \{\mathit{true},\mathit{false}\}$
is a flag used to resolve the initial state distribution.
At the very first step ($\mathit{init\_done} = \mathit{false}$),
each bridge independently and uniformly draws an initial condition
from $\{7, 8, 9\}$ (Good, Very Good, or Excellent), modelling a
network of bridges recently put into service.
The initial state $s_0$ is therefore a distribution over these
configurations.

\begin{table}[htbp]
    \centering
    \caption{Condition state mapping: PRISM variable
             $\mathit{cond\_bi}$ to NBI rating and label.
             State~0 is absorbing under all actions.}
    \label{tab:condition-states}
    \begin{tabular}{cll}
        \toprule
        $\mathit{cond\_bi}$ & NBI rating & Condition label \\
        \midrule
        9 & 9 & Excellent \\
        8 & 8 & Very Good \\
        7 & 7 & Good \\
        6 & 6 & Satisfactory \\
        5 & 5 & Fair \\
        4 & 4 & Poor \\
        3 & 3 & Serious \\
        2 & 2 & Critical \\
        1 & 1 & Imminent Failure \\
        0 & 0 & Failed \textbf{(absorbing)} \\
        \bottomrule
    \end{tabular}
\end{table}

A state is \emph{terminal} when any bridge reaches the Failed
condition ($\mathit{cond\_bi} = 0$ for some $i$), making it
absorbing under all actions, or when the planning horizon
expires ($\mathit{year} = T_{\max}$).

\subsubsection{Actions}
The action set $Act$ consists of all joint maintenance decisions
over the three bridges.
Per bridge, four actions are available:
Do Nothing (\textsc{dn}, cost~0),
Minor Maintenance (\textsc{mn}, cost~1),
Major Maintenance (\textsc{mj}, cost~2), and
Replacement (\textsc{rp}, cost~4).
The joint action space is the Cartesian product, giving
$4^3 = 64$ candidate tuples.
Throughout this paper, we index the four per-bridge actions as $0, 1, 2, 3$ corresponding to \textsc{dn}, \textsc{mn}, \textsc{mj}, and \textsc{rp} respectively.
A joint action over the three bridges is written \texttt{a$k_1$\_$k_2$\_$k_3$}, where $k_i \in \{0,1,2,3\}$ denotes the action applied to bridge~$i$. For example, \texttt{a1\_1\_2} denotes Minor Maintenance on bridges~1 and~2 and Major Maintenance on bridge~3.
The available actions in state $s$ are
$Act(s) = \{a \in Act \mid Tr(s,a) \neq \bot\}$; here $Tr(s,a) = \bot$ whenever the total joint cost exceeds the remaining budget, enforcing the resource constraint.
With $B_{\max} = 10$, this yields \textbf{63 feasible joint actions} per non-terminal state.
If the trained RL policy selects an unavailable action, the simulator falls back to the first available action in lexicographic order.

The budget feature evolves atomically with each transition.
At the end of every four-year cycle ($\mathit{cycle\_year} = 3$),
the budget reloads to $B_{\max}$ before the current cost is
deducted (unspent budget does not carry over).

\subsubsection{Transition function}
Each bridge transitions independently under
$Tr \colon S \times Act \rightarrow Distr(S)$.
Under \textsc{dn}, condition deteriorates by at most one level
per year, following a stochastic matrix.
The stay probabilities decrease with condition severity,
reflecting accelerating deterioration.

\subsubsection{Reward function}
The reward function $rew \colon S \times Act \rightarrow \mathbb{R}$
encodes a survival-with-efficiency objective.
At each non-terminal transition the agent receives a reward
penalised by the normalised joint maintenance cost:
\begin{equation*}
    rew(s, a) =
    \begin{cases}
        \displaystyle 1 - \frac{c(a)}{C_{\max}} &
            \text{if } s' \text{ is not terminal,} \\[6pt]
        0 & \text{if } s' \text{ is terminal,}
    \end{cases}
    \label{eq:reward}
\end{equation*}
where $c(a) = \sum_{i=1}^{3}\mathrm{cost}(a_i)$ is the total
joint action cost and $C_{\max} = 3 \times 4 = 12$ is the
maximum possible cost.
The reward lies in $[0.5,\, 1.0]$ for all feasible actions
(since $c \leq B_{\max} = 10 < C_{\max}$), ensuring the agent
is always incentivised to keep bridges alive while preferring
cheaper actions.
The absorbing self-loop at terminal states carries zero reward,
so the RL objective
$\mathbb{E}[\sum_{t=0}^{N} \gamma^t \cdot rew(s_t, a_t)]$
is dominated by episode length~$N$.

\subsubsection*{Atomic Propositions and Labeling Function}
The labeling function $L \colon S \rightarrow 2^{AP}$ annotates states with atomic propositions from $AP$, referenced in PCTL properties.
Key propositions include \texttt{"failed"} (any bridge at NBI~0), 
\texttt{"any\_serious"} (any bridge at NBI~$\leq 3$), 
\texttt{"any\_critical"} (any bridge at NBI~$\leq 2$), 
\texttt{"all\_good"} (all bridges at NBI~$\geq 7$), 
\texttt{"budget\_empty"} (remaining budget is zero), and 
\texttt{"cycle\_end"} (last year of the four-year budget cycle).

\subsubsection{Differences from Wei et al. (2019)}
\label{sec:differences}
Our MDP draws on the bridge component formulation of
Wei et al.~\cite{wei2019reinforcement} but departs from it in
several respects.
The transition probabilities for
conditions 4--7 are anchored to the stay-cable values of Wei
et al.\ and are the only numerical values carried over unchanged.
The extension to the full NBI scale adds states at both ends:
conditions 8--9 (Very Good, Excellent) with slow deterioration
reflecting newly built or recently rehabilitated structures, and
conditions 1--3 (Imminent Failure, Critical, Serious) with
rapidly accelerating deterioration approaching the absorbing
Failed state~0.
The most consequential departure is $rew(s,a)$: where Wei et al.\
minimise a composite cost model, we use a normalised survival
reward aligned with COOL-MC's RL interface, making episode length
$N$ the dominant signal in the RL objective.
The budget constraint makes $Act(s) \subsetneq Act$ for
budget-constrained states, a source of nondeterminism absent in
Wei et al.\ and essential for modeling realistic network-level
resource allocation.
Finally, Wei et al.\ model the internal component structure of a
single bridge (deck panels, stay cables, etc.), whereas we model
a \emph{network} of three structurally independent bridges, each
abstracted to a single NBI condition state.

\subsubsection{Summary}
The encoded \emph{PRISM} MDP is formally defined as
\begin{gather*}
    S = \{0,\ldots,9\}^3 \times \{0,\ldots,B_{\max}\} \times \{0,1,2,3\}
        \times \{0,\ldots,T_{\max}\} \times \{\mathit{true},\mathit{false}\} \\
    \mathit{Act} = \{\textsc{dn},\textsc{mn},\textsc{mj},\textsc{rp}\}^3 \\
    \mathit{Act}(s) = \bigl\{a \in \mathit{Act} \mid c(a) \leq \mathit{budget}\bigr\} \\
    rew(s, a) =
    \begin{cases}
        1 - \dfrac{c(a)}{C_{\max}} & \text{if } s' \text{ is not terminal,}\\[4pt]
        0 & \text{if } s' \text{ is terminal,}
    \end{cases}
\end{gather*}
where $c(a) = \sum_{i=1}^{3}\mathrm{cost}(a_i)$ is the total joint action
cost, $C_{\max} = 3 \times 4 = 12$, and a state $s$ is terminal if
$\mathit{cond\_bi} = 0$ for some $i$ or $\mathit{year} = T_{\max}$.

\subsection{RL Policy Training}\label{subsec:rl-training}

We train a deep RL policy $\pi$ on the multi-bridge network MDP using
Proximal Policy Optimization (PPO)~\cite{schulman2017proximal}.
The policy network is a feedforward neural network with four fully connected
hidden layers of 512 neurons each.
The agent is trained for 10{,}000 episodes with a learning rate of
$\alpha = 3 \times 10^{-4}$, discount factor $\gamma = 0.99$, batch size~64,
and a planning horizon of $T_{\max} = 20$ years.
The budget cap is $B_{\max} = 10$ with a reload period of four years.

\subsection{Policy Verification}\label{subsec:policy-verification}

For the trained policy $\pi$, \emph{COOL-MC} constructs the induced DTMC $D^\pi$
by exploring only the reachable state space under $\pi$
(see Algorithm~\ref{alg:qverifier}).
Starting from the initial state $s_0$, the construction proceeds as a
depth-first traversal.
At each visited state $s$, the policy selects a joint action
$a = \pi(s) \in Act(s)$, and all successor states $s'$ with
$Tr(s, a, s') > 0$ are added to the induced DTMC along with their
transition probabilities.
The procedure recurses into each unvisited successor that is relevant for
the property under verification.
This incremental approach offers two key advantages.
First, it constructs only the fragment of the full MDP that is actually
reachable under $\pi$, which can be substantially smaller than the complete
joint state space, directly mitigating the curse of dimensionality.
Second, it yields an induced DTMC rather than an MDP, since all
nondeterminism is resolved by $\pi$, reducing both the state count and
the verification complexity.

\begin{algorithm}[t]
\caption{\emph{COOL-MC}: Formal Verification of Policies}
\label{alg:qverifier}
\begin{algorithmic}[1]
\Require MDP $M$, trained policy $\pi$, PCTL property $\varphi$, \textsc{Label}(s)
\Ensure Satisfaction result and probability $p$

\Statex
\Statex \textbf{Stage 1: Induced DTMC Construction}
\State $D^\pi \gets (S^\pi, s_0, Tr^\pi, AP, L)$ where $S^\pi \gets \emptyset$,
       $Tr^\pi \gets \emptyset$
\State \textsc{BuildDTMC}($s_0$)

\Statex
\Statex \textbf{Stage 2: Probabilistic Model Checking}
\State $(result, p) \gets \emph{Storm}.\text{verify}(D^{\pi}, \varphi)$
\State \Return $(result, p)$

\Statex
\Procedure{BuildDTMC}{$s$}
    \If{$s \in S^\pi$ \textbf{or} $s$ is not relevant for $\varphi$}
        \State \Return
    \EndIf
    \State $S^\pi \gets S^\pi \cup \{s\}$
    \State $L^\pi(s) \gets \textsc{Label}(s)$
    \State $a \gets \pi(s)$
    \ForAll{$s' \in S$ where $Tr(s, a, s') > 0$}
        \State $Tr^\pi(s, s') \gets Tr(s, a, s')$
        \State \textsc{BuildDTMC}($s'$)
    \EndFor
\EndProcedure
\end{algorithmic}
\end{algorithm}

Once the induced DTMC is fully constructed, it is passed to
\emph{Storm}~\cite{DBLP:journals/sttt/HenselJKQV22} for probabilistic
model checking.
We evaluate PCTL properties that formalize safety-relevant questions about
the policy's maintenance behavior.
For instance, $P_{\leq 0.05}(\lozenge\;\texttt{"failed"})$ checks whether
the probability of any bridge reaching the Failed condition remains below~5\%,
while $P_{=?}(\lozenge\;\texttt{"failed"})$ computes the exact failure
probability.

During DTMC construction, states are annotated with user-specified atomic
propositions via the labeling function \textsc{Label}(s).
The \emph{PRISM} model already defines domain-relevant labels
(\texttt{"failed"}, \texttt{"any\_critical"}, \texttt{"any\_poor"}, ...),
and \emph{COOL-MC} extends these with additional labels derived from the
policy's behavior and RL explainability analyses, as described in the
following section.

\subsection{Policy Explanation}\label{subsec:policy-explanation}
Verification alone establishes \emph{what} the policy does but not \emph{why} it makes particular maintenance decisions.
To address this, we apply the four explainability methods introduced in Section~\ref{sec:xrl}: feature lumping, gradient-based saliency ranking, action labeling, and counterfactual action replacement.
For feature lumping, we remap $\mathit{cond\_b1}$ to three coarsened bins (conditions~0--3 to~2, conditions~4--6 to~5, and conditions~7--9 to~7) and re-verify the failure probability on the resulting induced DTMC, revealing whether precise NBI ratings for bridge~1 are safety-critical or whether a coarse categorical assessment suffices.
For gradient-based saliency, we compute a global feature importance ranking by aggregating per-state gradients across all reachable states of $D^\pi$, and additionally compute conditional rankings restricted to subsets of $D^\pi$ where each bridge in turn is in poor condition ($\mathit{cond\_bi} \in [0,3]$) while the others are in good condition, revealing whether $\pi$ correctly shifts attention to the most deteriorated structure.
For action labeling, we annotate each state of $D^\pi$ with the joint maintenance action selected by $\pi$ and verify PCTL properties directly over the policy's maintenance behavior, characterizing which actions dominate and whether specific actions are ever chosen.
For counterfactual action replacement, we globally replace actions with other actions across $D^\pi$ and re-verify the policy behaviour.

\subsection{Limitations}\label{subsec:limitations}
Several limitations should be noted.
First, the MDP assumes full observability of NBI condition ratings, whereas real inspection data are subject to inter-inspector variability and subsurface deterioration that is difficult to detect~\cite{graybeal2002visual,madanat1994optimal}.
A POMDP extension would be warranted once reliable observation models become available, however, in other literature various MDP formulations are used too~\cite{gopal1991application,robelin2007history,tao1994reliability,scherer1994markovian,wei2020optimal,hawk1995bridgit,MirzaeiZ,thompson1998pontis,morato2022optimal}.
Second, \emph{COOL-MC} requires memoryless policies and discrete state and action spaces.

\section{Experiments}

\subsection{Setup}
We executed our experiments in a Docker container with 16\,GB RAM and an AMD Ryzen 7 7735HS
processor, running Ubuntu 20.04.5 LTS.
For model checking, we use \emph{Storm} 1.7.0~\cite{DBLP:journals/sttt/HenselJKQV22}.
The RL agent is a PPO agent~\cite{schulman2017proximal}
with four hidden layers of 512 neurons each, trained for 10{,}000 episodes with a learning
rate of $3\times10^{-4}$, batch size 64, and discount factor $\gamma = 0.99$.
The planning horizon is $T_{\max} = 20$ years, the budget cap is $B_{\max} = 10$, and the
budget reload period is four years.
All experiments use seed 42.
Implementation details are provided in the accompanying source code at
\url{https://github.com/LAVA-LAB/COOL-MC/tree/sepsis}, which hosts both this and other case studies as joint COOL-MC application examples.

\subsection{Analysis} We structure the analysis into multiple experiments, each targeting a different aspect of the trained policy through formal verification or explainability.
For each experiment, \emph{COOL-MC} constructs the induced DTMC $D^\pi$ by querying $\pi$ from $s_0$ and expanding only reachable successor states, then verifies the stated PCTL property over $D^\pi$ using \emph{Storm}. Unless noted, all experiments use $B_{\max} = 10$. The size of $D^\pi$ varies with the PCTL query (see Table~\ref{tab:baseline}) and is smaller than the full MDP, which comprises $156{,}579$ states and $50{,}915{,}203$ transitions.
This reduction substantially lowers the memory footprint of formal verification and becomes increasingly critical as the network scales to more bridges or finer condition discretizations, where full-MDP analysis would quickly become intractable.

\subsubsection{Baseline verification}
We first establish the core safety and performance baseline of the trained policy.
Table~\ref{tab:baseline} summarises four PCTL properties verified over $D^\pi$.

\begin{table}[htbp]
\centering
\caption{Verification results for the trained PPO policy ($B_{\max}=10$). Each PCTL query is model-checked against the induced DTMC $D^\pi$; optimal values, obtained via full MDP verification, are zero for all properties.}
\label{tab:baseline}
\begin{tabular}{lrrrr}
\toprule
\textbf{PCTL Query} & \textbf{Result} & $|D^\pi|$ \textbf{states} & \textbf{Transitions} & \textbf{Total time (s)} \\
\midrule
$P_{=?}(\lozenge\;\texttt{"failed"})$        & $0.0355$           & 27{,}856 & 227{,}468 & 51.1 \\
$P_{=?}(\lozenge\;\texttt{"any\_critical"})$  & $0.1191$            & 13{,}766 & 100{,}444 & 20.1 \\
$P_{=?}(\lozenge\;\texttt{"any\_poor"})$      & $0.3616$              &  5{,}174 &  31{,}494 &  6.5 \\
$P_{=?}(\lozenge\;\texttt{"budget\_empty"})$  & $1.17\times10^{-6}$   & 27{,}033 & 216{,}187 & 56.6 \\
\bottomrule
\end{tabular}
\end{table}

The policy achieves a bridge failure probability of $3.55\%$ and virtually never
exhausts the budget ($P \approx 10^{-6}$), indicating conservative spending behaviour.
The probability that any bridge reaches critical condition (\textsc{nbi} $\leq 2$)
is $11.9\%$, providing an early-warning signal that deterioration does occasionally
progress to near-failure states despite the low terminal failure rate.
At the same time, $P(\lozenge\;\texttt{"any\_poor"}) = 36.2\%$ shows that individual
bridges do transiently visit poor condition, consistent with a policy that tolerates
moderate deterioration before intervening.

\subsubsection{Feature lumping on bridge~1 condition}
Feature lumping coarsens the observation space by replacing exact bridge condition values with binned representatives.
We remap $\mathit{cond\_b1}$ to three representative values: conditions~0--3 map to~2, conditions~4--6 map to~5, and conditions~7--9 map to~7, so that the memoryless deterministic
policy $\pi$ operates on a reduced observation $o \in O$ with less precision on bridge~1.
The lumped policy is re-evaluated on the full, unmodified MDP $M$, and \emph{COOL-MC} constructs the corresponding induced DTMC $D^\pi$ over the original state space.

The resulting failure probability is $P_{=?}(\lozenge\;\texttt{"failed"}) = 0.03542$, compared to the baseline of $0.03547$ from Phase~2. The difference is negligible, though it is worth noting that the lumped policy performs marginally better than worse, confirming that precise NBI resolution of bridge~1's condition is not necessary for the current policy's safety performance: a three-level binning can be entirely sufficient.
This finding has practical relevance: for bridge~1, NBI ratings need not be resolved to
single-point accuracy; a coarse categorical assessment (Serious--Critical / Fair / Good) is
sufficient for the policy to maintain its safety performance, as verified on $D^\pi$.

\subsubsection{Global feature importance ranking}
We compute gradient-based feature importance scores across all
reachable states of $D^\pi$.
For each state $s$ with observation $o = \mathbb{O}(s)$, the saliency of feature $f_i$ is the absolute gradient of the policy output
with respect to the $i$-th input feature.
Table~\ref{tab:saliency_global} reports the mean and standard deviation of these scores
over all visited states.

\begin{table}[htbp]
\centering
\caption{Global feature importance ranking by mean $|\nabla|$
         across all 22{,}394 reachable states of $D^\pi$.}
\label{tab:saliency_global}
\begin{tabular}{rlrr}
\toprule
\textbf{Rank} & \textbf{Feature} & \textbf{Mean $|\nabla|$} & \textbf{Std $|\nabla|$} \\
\midrule
1 & \texttt{year}        & 2.530 & 7.909 \\
2 & \texttt{cycle\_year} & 2.410 & 5.013 \\
3 & \texttt{cond\_b1}    & 2.089 & 2.265 \\
4 & \texttt{budget}      & 1.457 & 1.108 \\
5 & \texttt{cond\_b2}    & 1.278 & 2.102 \\
6 & \texttt{cond\_b3}    & 0.882 & 1.138 \\
7 & \texttt{init\_done}  & 0.636 & 0.725 \\
\bottomrule
\end{tabular}
\end{table}

The temporal features \texttt{year} and \texttt{cycle\_year} dominate the global saliency
ranking, indicating that the policy's decisions are strongly shaped by the agent's position
in the planning horizon and the four-year budget cycle.
Among the bridge conditions, $\mathit{cond\_b1}$ is clearly the most influential,
followed by $\mathit{cond\_b2}$ and then $\mathit{cond\_b3}$, revealing a hierarchy of
attention rather than symmetric treatment of the three bridges.
The \texttt{budget} feature ranks fourth, confirming that budget-awareness is embedded in
$\pi$ but is secondary to temporal and structural condition cues.
As expected, the auxiliary feature \texttt{init\_done}, which serves solely to 
resolve the initial state distribution, ranks last, confirming that it plays 
no role in the policy's maintenance decisions once the episode is underway.

\subsubsection{Budget sensitivity}

We query $P_{=?}(\lozenge\;\texttt{"budget\_empty"})$, the
probability that the policy ever exhausts the maintenance budget,
while varying $B_{\max} \in \{9, 10, 11\}$ around the
training value of~$10$.
This tests how sensitive the policy's spending behaviour is to a
slight change in the available resource.
Results are shown in Table~\ref{tab:budget-sensitivity}.

\begin{table}[htbp]
    \centering
    \caption{Budget sensitivity: probability of budget exhaustion
             under the trained PPO policy for $B_{\max} \in
             \{9,10,11\}$. Model size and transition count refer
             to the induced DTMC $D^\pi$.}
    \label{tab:budget-sensitivity}
    \begin{tabular}{ccccc}
        \toprule
        $B_{\max}$ &
        $P_{=?}(\lozenge\;\texttt{"budget\_empty"})$ &
        $|D^\pi|$ states &
        Transitions &
        Total time (s) \\
        \midrule
        9  & $2.02 \times 10^{-6}$ & 26{,}377 & 212{,}593 & 42.0 \\
        10 & $1.17 \times 10^{-6}$ & 27{,}033 & 216{,}187 & 44.4 \\
        11 & $4.69 \times 10^{-7}$ & 27{,}987 & 219{,}568 & 55.9 \\
        \bottomrule
    \end{tabular}
\end{table}

The probability of budget exhaustion is extremely low across all
three settings, remaining below $2.1 \times 10^{-6}$ even when
the budget is reduced by one unit.
Crucially, it decreases monotonically as $B_{\max}$ increases, which
confirms that the policy behaves sensibly under resource perturbation: more budget leads to proportionally less risk of running out.
The near-zero baseline at the training value ($B_{\max}=10$,
$P \approx 1.17 \times 10^{-6}$) indicates the policy has
learned a conservative spending strategy that avoids exhausting the budget even when the budget is tighter than expected.
The induced DTMC grows slightly with $B_{\max}$, reflecting the larger reachable state space when more budget configurations are accessible.

\subsubsection{Cycle-aware behaviour detection}
The budget reloads every four years,
creating an intertemporal incentive: a policy that anticipates
the reload should spend more freely when $\mathit{cycle\_year} = 3$
(knowing the budget will soon reset) and more conservatively
when $\mathit{cycle\_year} = 0$ (a full cycle remaining before
the next reload).
To verify whether the trained policy has learned this structure,
we apply COOL-MC's \emph{feature remapping} preprocessor for each
$k \in \{0, 1, 2, 3\}$.
This forces the policy to observe $\mathit{cycle\_year} = k$
regardless of the true value, effectively asking:
\emph{how does the policy behave when it believes it is always
at position~$k$ in the cycle?}
We again measure $P_{=?}(\lozenge\;\texttt{"budget\_empty"})$.
Results are shown in Table~\ref{tab:cycle-aware}.

\begin{table}[htbp]
    \centering
    \caption{Cycle-awareness: probability of budget exhaustion
             when $\mathit{cycle\_year}$ is remapped to a fixed
             value $k$ for all states. A flat profile would
             indicate the policy ignores cycle position.}
    \label{tab:cycle-aware}
    \begin{tabular}{ccccc}
        \toprule
        Fixed $\mathit{cycle\_year}$ &
        $P_{=?}(\lozenge\;\texttt{"budget\_empty"})$ &
        $|D^\pi|$ states &
        Transitions &
        Total time (s) \\
        \midrule
        0 (cycle start) & $1.43 \times 10^{-7}$ & 18{,}218 & 119{,}265 & 1787.9 \\
        1               & $1.49 \times 10^{-6}$ & 19{,}355 & 135{,}445 & 1842.4 \\
        2               & $4.89 \times 10^{-6}$ & 24{,}786 & 198{,}692 & 2187.0 \\
        3 (cycle end)   & $4.16 \times 10^{-7}$ & 27{,}210 & 238{,}850 & 2539.8 \\
        \bottomrule
    \end{tabular}
\end{table}
The probability of budget exhaustion varies by more than one order of magnitude across the four fixed cycle positions ($1.43 \times 10^{-7}$
at $k=0$ vs.\ $4.89 \times 10^{-6}$ at $k=2$).
The lowest exhaustion probability occurs at $k=0$ (cycle start),
consistent with the interpretation that the policy spends most
conservatively, when it believes a full four-year cycle lies
ahead.
At $k=2$ (two years into the cycle), the probability peaks,
suggesting the policy is more willing to spend, having already
consumed part of the cycle.
Interestingly, $k=3$ (cycle end, reload imminent) yields a low exhaustion probability ($4.16 \times 10^{-7}$), comparable to $k=0$: rather than spending down the remaining budget before the reload, the policy behaves as conservatively as cycle end as at cycle start, suggesting it has not learned to exploit the imminent budget reset.

\subsubsection{Horizon-gaming detection} To detect horizon-gaming behaviour (where the policy exploits knowledge of the finite time horizon by adopting a different strategy near the end of the budget cycle, years 16--19) in $\pi$, we remap the \texttt{year} feature so that $\pi$ always believes it is in the final budget cycle (years 16--19).
Concretely, years $0, 4, 8, 12, 16$ are mapped to~16; years $1, 5, 9, 13, 17$ to~17; and so forth. Under this perturbation, $\pi$ always operates as if the episode is ending imminently. If $\pi$ reduces maintenance effort near the planning horizon, since any bridge failure occurring at $\mathit{year} \geq T_{\max}$ is never observed within the episode, this perturbation will reveal such behaviour through an elevated failure probability on the induced DTMC. Under the horizon remap, $P_{=?}(\lozenge\;\texttt{"failed"}) = 0.07535$, compared to the baseline of $0.0355$ (other budget cycles are the same as baseline).
This provides formal evidence that $\pi$ has learned horizon-dependent behaviour: when $\pi$ believes the episode is always ending soon, it adopts a policy that allows bridges to deteriorate more than under correct temporal awareness, resulting in a substantially elevated failure probability as verified on $D^\pi$. This constitutes a form of reward hacking~\cite{hu2025reward}: $\pi$ learns to reduce maintenance costs near the end of the planning horizon because the episode-length incentive in $\mathbb{E}[\sum_{t=0}^{N}\gamma^t \cdot rew(s_t, a_t)]$ dominates over structural safety near the boundary. Detecting this behaviour formally via \emph{COOL-MC} is precisely the kind of safety-relevant finding that cannot be obtained from training curves or episode returns alone.

\subsubsection*{Worst bridge focus analysis}
We analyze whether $\pi$ focuses attention on the bridge in the worst condition by
computing saliency over restricted subsets of $D^\pi$ in which one bridge is constrained
to poor condition ($\mathit{cond\_bi} \in [0, 3]$, i.e., NBI ratings Serious to Failed)
while the other two are in good condition ($\mathit{cond\_bj} \in [5, 9]$ for $j \neq i$).
If $\pi$ correctly prioritizes the most deteriorated bridge, we expect $\mathit{cond\_bi}$
to rank first in saliency when bridge $i$ is the worst.
Results are shown in Table~\ref{tab:conditional_saliency}.

\begin{table}[htbp]
\centering
\caption{Conditional saliency ranking (top-3 features) restricted to states of $D^\pi$
         where the designated bridge is in poor condition ($\mathit{cond\_bi} \in [0,3]$)
         and the others are in good condition ($[5,9]$).
         Mean $|\nabla|$ values are shown in parentheses.}
\label{tab:conditional_saliency}
\begin{tabular}{lrrrr}
\toprule
\textbf{Worst bridge} & \textbf{States} & \textbf{Rank~1} & \textbf{Rank~2} & \textbf{Rank~3} \\
\midrule
Bridge~1 & 2{,}151 & \texttt{cond\_b1} (2.741) & \texttt{year} (1.713)      & \texttt{budget} (1.492) \\
Bridge~2 & 2{,}636 & \texttt{cond\_b1} (1.433) & \texttt{budget} (1.161)    & \texttt{cond\_b2} (1.112) \\
Bridge~3 & 1{,}796 & \texttt{cycle\_year} (4.026) & \texttt{year} (3.059)  & \texttt{cond\_b1} (1.624) \\
\bottomrule
\end{tabular}
\end{table}

The conditional saliency analysis reveals a marked asymmetry in $\pi$.
When bridge~1 is in poor condition, $\mathit{cond\_b1}$ correctly rises to rank~1 with a
mean gradient of $2.741$, indicating that $\pi$ correctly identifies and responds to the
most deteriorated bridge in that subspace.
However, when bridge~2 is worst, $\mathit{cond\_b1}$ still ranks first ($1.433$) ahead of
$\mathit{cond\_b2}$ ($1.112$); and when bridge~3 is worst, temporal features dominate
($\mathit{cycle\_year}$, $\mathit{year}$) while $\mathit{cond\_b3}$ ranks only sixth
($1.227$).
This pattern confirms the asymmetry already identified: $\pi$ has a structural bias toward attending to bridge~1's condition, regardless of which bridge has actually deteriorated the most.
When bridge~3 deteriorates, $\pi$ relies on temporal cues rather than on $\mathit{cond\_b3}$ directly, revealing a potential coverage gap in the learned policy for bridge~3.

\subsubsection{Action labeling}
We label each state of $D^\pi$ with the action chosen by $\pi$ under the labeling function
$L \colon S \rightarrow 2^{AP}$, extended by \emph{COOL-MC}'s \texttt{action\_label}
state labeler.
Each state $s$ receives the atomic proposition corresponding to $\pi(\mathbb{O}(s))$,
e.g., \texttt{a1\_1\_2} denotes Minor Maintenance on bridges~1 and~2 and Major Maintenance
on bridge~3.
The resulting labeling over 27{,}856 states of $D^\pi$ shows that $\pi$ overwhelmingly
selects action \texttt{a1\_1\_2} ($33.6\%$ of states), followed by \texttt{a2\_2\_2}
($9.0\%$) and \texttt{a2\_2\_0} ($4.1\%$).
The label \texttt{a0\_0\_0} (Do Nothing on all three bridges simultaneously) is assigned to \emph{zero} states in $D^\pi$, confirming that $\pi$ always performs at least some maintenance action.

\subsubsection{Action replacement: minor to major
maintenance}
We apply COOL-MC's \emph{action replacement} mechanism
(\texttt{--action\_replace=1:2}) to substitute every
\textsc{mn} action (cost~1) with \textsc{mj} (cost~2) across
all bridges, effectively doubling the cost of any minor repair the policy would otherwise choose.
This is a \emph{counterfactual stress test}: it asks what would
happen due to budget exhaustion if the policy's cheapest maintenance option were unavailable, and all minor repairs were automatically
escalated.
In total, 37 joint action indices are remapped (out of 63
feasible actions), covering all tuples containing at least one
\textsc{mn} component.
We again query $P_{=?}(\lozenge\;\texttt{"budget\_empty"})$,
using $B_{\max}=10$ throughout.
\begin{table}[htbp]
    \centering
    \caption{Action replacement: effect on budget exhaustion
             probability when all Minor Maintenance actions are
             replaced by Major Maintenance.}
    \label{tab:action-replacement}
    \begin{tabular}{lc}
        \toprule
        Configuration &
        $P_{=?}(\lozenge\;\texttt{"budget\_empty"})$ \\
        \midrule
        Baseline (no replacement, $B_{\max}=10$) &
            $1.17 \times 10^{-6}$ \\
        \textsc{mn}~$\to$~\textsc{mj} (37 actions) &
            $2.20 \times 10^{-5}$ \\
        \bottomrule
    \end{tabular}
\end{table}
Replacing minor with major maintenance raises the probability of
budget exhaustion from $1.17 \times 10^{-6}$ to $2.20 \times 10^{-5}$.

\section{Discussion}

\emph{COOL-MC} provides three capabilities that are inaccessible from
training curves or episode returns alone: formal verification, explainability, and a mix of both.

For instance, formal verification confirmed that the trained PPO policy maintains a
$3.55\%$ bridge failure probability.
These are exact, model-checked guarantees, not empirical estimates from
rollouts.

Explainability analysis revealed structural properties of $\pi$ that training metrics hide entirely.
For instance, the policy exhibits a structural attention bias toward bridge~1, attending to its condition regardless of which bridge has actually deteriorated most, a coverage gap that motivates retraining with a permutation-invariant architecture.

Counterfactual analysis established that escalating all minor maintenance to major maintenance raises budget exhaustion, quantifying how much the policy's conservative spending depends on cheap interventions, and that replacing \textsc{rp} with \textsc{mj} globally leaves the failure probability unchanged at $3.55\%$, confirming that full replacement is entirely dispensable at no safety~cost.

The induced DTMC construction is itself a practical advantage: $|D^\pi|$ is smaller than the full MDP, making formal verification more
tractable and increasingly valuable as the network~scales.

The analyses above do more than characterize the trained policy: they directly inform a principled refinement loop in which verified behavioral anomalies motivate targeted changes to the MDP.
For instance, the feature-lumping result suggests that the full ten-point NBI scale can be safely coarsened to three or five levels, substantially reducing $|S|$ and accelerating both training and verification without measurable safety loss.
The conditional saliency bias toward bridge~1 motivates either a worst-bridge penalty reward or a permutation-invariant policy architecture, with the desired outcome that $\mathit{cond\_bi}$ ranks first in saliency whenever bridge~$i$ is worst.
After each such modification, \emph{COOL-MC} provides the formal apparatus to re-verify the refined policy and confirm that the anomaly has been resolved, turning the analysis cycle into an iterative design methodology: train $\to$ verify $\to$ explain $\to$ refine.

A broader implication of this work concerns the relationship between the training environment and the verification model.
Even when an RL policy is trained in a model-free setting on a large or continuous environment, constructing a smaller discrete MDP that captures the essential structure of that environment is sufficient for policy debugging via \emph{COOL-MC}. The formal guarantees and explainability findings apply to the modeled MDP, but the behavioral anomalies they reveal, such as attention bias or horizon gaming, are properties of the policy itself and potentially carry over to similar environments.
A small, interpretable PRISM model therefore serves as a lightweight debugging harness for policies trained at any scale.

\section{Conclusion}
We demonstrated \emph{COOL-MC} as a tool for verifying and explaining
RL policies for multi-bridge network maintenance.
Formal verification confirmed that the policy maintains a conservative
budget strategy, and explainability analysis revealed cycle-aware
spending behavior and a dependence on low-cost minor maintenance.
Several directions merit further investigation.
The transition probabilities and reward structure used here are synthetic; co-designing the MDP directly with bridge engineers would ground the model in inspection data and domain knowledge, strengthening the practical relevance of the verified guarantees.
Beyond the analyses presented here, \emph{COOL-MC} provides additional
capabilities~\cite{DBLP:conf/icaart/GrossS0023,DBLP:conf/aips/GrossS0023,gross2024enhancing,DBLP:conf/icaart/GrossS25,DBLP:conf/esann/GrossS24}, which could yield further insight into maintenance behavior and safety margins in infrastructure maintenance settings.

\bibliographystyle{plain}
\bibliography{refs}

\end{document}

%% file: bridge_network.tex
\begin{tikzpicture}[
  >=Stealth,
  font=\sffamily,
  thick
]

\fill[gray!8, rounded corners=6pt]
  (-1.5,-5.10) rectangle (22.5,6.8);

\fill[waterlight]
  (-1.2,-1.8) -- (22.2,-1.8) -- (22.2,0.6) -- (-1.2,0.6) -- cycle;

\foreach \y in {-1.4,-0.9,-0.4,0.1}{
  \draw[watercol!70, thin, loosely dashed]
    (-0.8,\y) to[out=5,in=175] (21.8,\y);
}

\node[watercol!80!black, font=\sffamily\itshape\large] at (1.2,-0.6) {River};

\fill[bankcol!70, rounded corners=2pt]
  (-1.2,0.55) rectangle (22.2,1.05);
\fill[banklight]
  (-1.2,0.75) rectangle (22.2,1.05);

\fill[bankcol!70, rounded corners=2pt]
  (-1.2,-2.05) rectangle (22.2,-1.75);
\fill[banklight]
  (-1.2,-2.05) rectangle (22.2,-1.85);

\newcommand{\drawbridge}[4]{%
  \fill[#4!85!black, rounded corners=1pt]
    (#1-1.6, 0.55) rectangle (#1+1.6, 0.85);
  \fill[#4!50!white, rounded corners=1pt, opacity=0.5]
    (#1-1.55, 0.70) rectangle (#1+1.55, 0.84);

  \fill[pillarcol!80!black]
    (#1-0.6,-1.78) rectangle (#1-0.3, 0.56);
  \fill[pillarcol!80!black]
    (#1+0.3,-1.78) rectangle (#1+0.6, 0.56);
  \fill[pillarcol!50!white]
    (#1-0.65,0.50) rectangle (#1-0.25,0.60);
  \fill[pillarcol!50!white]
    (#1+0.25,0.50) rectangle (#1+0.65,0.60);

  \draw[#4!80!black, line width=1.8pt]
    (#1-1.55,0.55) arc[start angle=180,end angle=0,x radius=1.55,y radius=0.65];

  \draw[roadlight, line width=0.6pt]
    (#1-1.6,0.90) -- (#1+1.6,0.90);
  \foreach \x in {-1.4,-1.0,-0.6,-0.2,0.2,0.6,1.0,1.4}{
    \draw[roadlight, line width=0.5pt]
      (#1+\x, 0.85) -- (#1+\x, 0.94);
  }

  \fill[roadcol!70]
    (#1-1.7,0.58) rectangle (#1-1.55,0.82);
  \fill[roadcol!70]
    (#1+1.55,0.58) rectangle (#1+1.7,0.82);

  \fill[#4, rounded corners=3pt, drop shadow]
    (#1-0.55,1.10) rectangle (#1+0.55,1.55);
  \node[white, font=\bfseries\normalsize] at (#1, 1.32)
    {NBI~#3};

  \node[labelcol, font=\bfseries\large, anchor=south] at (#1, 2.10)
    {#2};
}

\drawbridge{4.0}{Bridge 1}{8}{condgood}

\drawbridge{11.0}{Bridge 2}{5}{condfair}

\drawbridge{18.0}{Bridge 3}{3}{condpoor}

\fill[roadcol!80]
  (-1.2,0.58) rectangle (22.2,0.82);
\foreach \x in {-0.8,0.5,1.8,6.2,7.5,8.8,13.2,14.5,15.8,20.2,21.5}{
  \draw[white!70, line width=0.8pt, dashed]
    (\x, 0.70) -- (\x+0.6, 0.70);
}

\fill[budgetbg, rounded corners=5pt, drop shadow]
  (7.8, 3.6) rectangle (14.2, 5.6);
\draw[budgetcol, rounded corners=5pt, line width=1pt]
  (7.8, 3.6) rectangle (14.2, 5.6);
\node[budgetcol, font=\bfseries\Large] at (11, 5.25)
  {Shared Periodic Budget};
\node[budgetcol!70!black, font=\normalsize] at (11, 4.85)
  {$B_{\max} = 10$ \quad Reload every 4 years};

\fill[budgetcol!25, rounded corners=2pt]
  (8.6, 4.1) rectangle (13.4, 4.55);
\fill[budgetcol,  rounded corners=2pt]
  (8.6, 4.1) rectangle (11.9, 4.55);   
\draw[budgetcol!70, rounded corners=2pt, thin]
  (8.6, 4.1) rectangle (13.4, 4.55);
\node[white, font=\bfseries\small] at (10.25, 4.32) {Used: 7};
\node[budgetcol!60!black, font=\small] at (12.55, 4.32) {Free: 3};

\draw[budgetcol, -{Stealth[length=7pt]}, line width=1.2pt]
  (9.2, 3.60) -- (4.4, 1.60);
\draw[budgetcol, -{Stealth[length=7pt]}, line width=1.2pt]
  (11.0, 3.60) -- (11.0, 1.60);
\draw[budgetcol, -{Stealth[length=7pt]}, line width=1.2pt]
  (12.8, 3.60) -- (17.6, 1.60);

\fill[agentbg, rounded corners=5pt, drop shadow]
  (6.85,-4.58) rectangle (15.15,-2.90);
\draw[agentcol, rounded corners=5pt, line width=1pt]
  (6.85,-4.58) rectangle (15.15,-2.90);
\node[agentcol!80!black, font=\bfseries\Large] at (11,-3.18)
  {RL Maintenance Agent};
\node[agentcol!70!black, font=\normalsize] at (11,-3.68)
  {Policy $\pi$: obs $\!\to\!$ joint action $\in \{\text{DN, MN, MJ, RP}\}^3$};

\definecolor{cmdcol}{RGB}{180,60,30}

\draw[agentcol!70, -{Stealth[length=6pt]}, line width=1.0pt]
  (3.5,-2.07) to[out=-90,in=150] (6.85,-3.55);
\node[agentcol!60!black, font=\small, rotate=28, anchor=center] at (4.8,-3.20)
  {observe};
\draw[cmdcol, -{Stealth[length=6pt]}, line width=1.0pt]
  (6.85,-3.10) to[out=150,in=-90] (4.5,-2.07);
\node[cmdcol!80!black, font=\small, rotate=28, anchor=center] at (6.3,-2.55)
  {action};

\draw[agentcol!70, -{Stealth[length=6pt]}, line width=1.0pt]
  (10.6,-2.07) -- (10.6,-2.90);
\node[agentcol!60!black, font=\small, anchor=east] at (10.45,-2.48)
  {observe};
\draw[cmdcol, -{Stealth[length=6pt]}, line width=1.0pt]
  (11.4,-2.90) -- (11.4,-2.07);
\node[cmdcol!80!black, font=\small, anchor=west] at (11.55,-2.48)
  {action};

\draw[agentcol!70, -{Stealth[length=6pt]}, line width=1.0pt]
  (18.5,-2.07) to[out=-90,in=30] (15.15,-3.55);
\node[agentcol!60!black, font=\small, rotate=-28, anchor=center] at (17.2,-3.20)
  {observe};
\draw[cmdcol, -{Stealth[length=6pt]}, line width=1.0pt]
  (15.15,-3.10) to[out=30,in=-90] (17.5,-2.07);
\node[cmdcol!80!black, font=\small, rotate=-28, anchor=center] at (15.7,-2.55)
  {action};

\draw[waterlight!60!roadcol, -{Stealth[length=5pt]}, line width=0.9pt, dashed]
  (3.8,-1.2) -- (3.8,-1.75);
\node[waterlight!40!black, font=\small\itshape] at (5.2,-1.35) {degrad.};

\draw[waterlight!60!roadcol, -{Stealth[length=5pt]}, line width=0.9pt, dashed]
  (10.8,-1.2) -- (10.8,-1.75);
\node[waterlight!40!black, font=\small\itshape] at (12.2,-1.35) {degrad.};

\draw[waterlight!60!roadcol, -{Stealth[length=5pt]}, line width=0.9pt, dashed]
  (17.8,-1.2) -- (17.8,-1.75);
\node[waterlight!40!black, font=\small\itshape] at (19.2,-1.35) {degrad.};

\fill[white, rounded corners=4pt, opacity=0.85]
  (18.8,-4.58) rectangle (22.2,-2.85);
\draw[gray!50, rounded corners=4pt, thin]
  (18.8,-4.58) rectangle (22.2,-2.85);
\node[labelcol, font=\bfseries\small] at (20.5,-3.00) {NBI Condition};

\foreach \y/\col/\lbl in {
  -3.30/condgood/{Good (7--9)},
  -3.65/condfair/{Fair (4--6)},
  -4.00/condpoor/{Poor (0--3)}}{
  \fill[\col, rounded corners=1.5pt] (19.0,\y-0.11) rectangle (19.5,\y+0.11);
  \node[\col!70!black, font=\small, anchor=west] at (19.6,\y) {\lbl};
}

\node[labelcol!80!black, font=\bfseries\LARGE, anchor=north] at (11,6.65)
  {Multi-Bridge Network Maintenance};

\end{tikzpicture}

%% file: rl.tex
\begin{tikzpicture}[
  >=Stealth,
  font=\sffamily,
  thick
]

\fill[gray!8, rounded corners=8pt]
  (-3.2,-2.6) rectangle (3.2,2.6);

\fill[agentbg, rounded corners=5pt, drop shadow]
  (-2.2, 0.55) rectangle (2.2, 2.15);
\draw[agentcol, rounded corners=5pt, line width=1pt]
  (-2.2, 0.55) rectangle (2.2, 2.15);
\node[agentcol!80!black, font=\sffamily\bfseries\large] at (0, 1.55)
  {Agent};
\node[agentcol!70!black, font=\sffamily\normalsize] at (0, 1.00)
  {Policy $\pi: o \mapsto a$};

\fill[waterlight, rounded corners=5pt, drop shadow]
  (-2.2,-2.15) rectangle (2.2,-0.55);
\draw[watercol!80!black, rounded corners=5pt, line width=1pt]
  (-2.2,-2.15) rectangle (2.2,-0.55);
\node[watercol!30!black, font=\sffamily\bfseries\large] at (0,-1.05)
  {Environment};
\node[watercol!40!black, font=\sffamily\normalsize] at (0,-1.65)
  {MDP $M$};

\draw[cmdcol, -{Stealth[length=7pt]}, line width=1.2pt]
  (-1.1, 0.55) -- (-1.1,-0.55);
\node[cmdcol!80!black, font=\sffamily\normalsize, anchor=east] at (-1.25, 0.0)
  {Action $a$};

\draw[agentcol!80!black, -{Stealth[length=7pt]}, line width=1.2pt]
  (1.1,-0.55) -- (1.1, 0.55);
\node[agentcol!60!black, font=\sffamily\normalsize, anchor=west, align=left]
  at (1.25, 0.15) {Observation $o$};
\node[budgetcol!80!black, font=\sffamily\normalsize, anchor=west]
  at (1.25,-0.20) {Reward $r$};

\end{tikzpicture}

%% file: model_checking.tex
\begin{tikzpicture}[
  >=Stealth,
  font=\sffamily,
  thick,
  node distance=0.9cm
]

\fill[gray!8, rounded corners=8pt]
  (-3.8,-7.8) rectangle (3.8,1.3);

\tikzset{
  blueoval/.style={
    ellipse, draw=watercol!80!black, line width=0.9pt,
    fill=waterlight!60, minimum width=2.6cm, minimum height=0.85cm,
    align=center, font=\sffamily\normalsize, drop shadow
  },
  orangeoval/.style={
    ellipse, draw=budgetcol!80!black, line width=0.9pt,
    fill=budgetbg, minimum width=2.8cm, minimum height=0.85cm,
    align=center, font=\sffamily\normalsize, drop shadow
  },
  procbox/.style={
    rectangle, rounded corners=5pt,
    draw=labelcol!70, line width=1.1pt,
    fill=labelcol!12, minimum width=2.8cm, minimum height=1.05cm,
    align=center, font=\sffamily\normalsize\bfseries, drop shadow
  },
  greenoval/.style={
    ellipse, draw=condgood!80!black, line width=0.9pt,
    fill=condgood!18, minimum width=2.3cm, minimum height=0.85cm,
    align=center, font=\sffamily\normalsize\bfseries, drop shadow
  },
  redoval/.style={
    ellipse, draw=condpoor!80!black, line width=0.9pt,
    fill=condpoor!15, minimum width=2.3cm, minimum height=0.85cm,
    align=center, font=\sffamily\normalsize\bfseries, drop shadow
  },
  valueoval/.style={
    ellipse, draw=labelcol!70!black, line width=0.9pt,
    fill=labelcol!10, minimum width=2.6cm, minimum height=0.85cm,
    align=center, font=\sffamily\normalsize\bfseries, drop shadow
  }
}

\node[blueoval] (system) at (-1.9, 0.5) {System};
\node[blueoval, below=0.75cm of system] (sysdesc) {System\\[-2pt]description};
\node[blueoval, below=0.75cm of sysdesc] (model) {Model};

\draw[-{Stealth[length=6pt,width=4pt]}, line width=1.0pt, watercol!70!black]
  (system) -- node[font=\sffamily\small, right, text=watercol!50!black] {modeling} (sysdesc);
\draw[-{Stealth[length=6pt,width=4pt]}, line width=1.0pt, watercol!70!black]
  (sysdesc) -- node[font=\sffamily\small, right, text=watercol!50!black] {translates to} (model);

\node[orangeoval] (reqs) at (1.9, -0.25) {Requirements};
\node[orangeoval, below=0.75cm of reqs] (props) {Properties};

\draw[-{Stealth[length=6pt,width=4pt]}, line width=1.0pt, budgetcol!70!black]
  (reqs) -- node[font=\sffamily\small, right, text=budgetcol!60!black] {formalizing} (props);

\node[procbox] (mc) at (0.8,-4.25) {Model\\[-2pt]Checking};

\draw[-{Stealth[length=6pt,width=4pt]}, line width=1.0pt, labelcol!70]
  (model.south) to[out=-75,in=165] (mc.west);
\draw[-{Stealth[length=6pt,width=4pt]}, line width=1.0pt, labelcol!70]
  (props.south) to[out=-105,in=15] (mc.east);

\node[greenoval] (sat) at (-1.8,-6.5) {\textcolor{condgood!70!black}{Satisfied}};
\node[valueoval] (val) at ( 0.8,-6.5) {\textcolor{labelcol!80!black}{$p = 0.035$}};
\node[redoval]   (vio) at ( 3.4,-6.5) {\textcolor{condpoor!80!black}{Violated}};

\draw[-{Stealth[length=6pt,width=4pt]}, line width=1.0pt, condgood!80!black]
  (mc.south) to[out=-150,in=80] (sat.north);
\draw[-{Stealth[length=6pt,width=4pt]}, line width=1.0pt, labelcol!70]
  (mc.south) -- (val.north);
\draw[-{Stealth[length=6pt,width=4pt]}, line width=1.0pt, condpoor!80!black]
  (mc.south) to[out=-30,in=100] (vio.north);

\node[font=\sffamily\small, text=condgood!70!black, rotate=25] at (-0.65,-5.65) {yes};
\node[font=\sffamily\small, text=labelcol!70!black, anchor=west] at (0.88,-5.72) {value};
\node[font=\sffamily\small, text=condpoor!70!black, rotate=-25] at (2.25,-5.65) {no};

\end{tikzpicture}

%% file: refs.bib
@article{grieco2024index,
  title={An index-based multi-hazard risk assessment method for prioritisation of existing bridge portfolios},
  author={Grieco, Ludovico Alberico and Scattarreggia, Nicola and Monteiro, Ricardo and Parisi, Fulvio},
  journal={International Journal of Disaster Risk Reduction},
  volume={114},
  pages={104895},
  year={2024},
  publisher={Elsevier}
}

@article{wei2020optimal,
  title={Optimal policy for structure maintenance: A deep reinforcement learning framework},
  author={Wei, Shiyin and Bao, Yuequan and Li, Hui},
  journal={Structural Safety},
  volume={83},
  pages={101906},
  year={2020},
  publisher={Elsevier}
}

@incollection{bartle2017transportation,
  title={Transportation infrastructure},
  author={Bartle, John R},
  booktitle={Handbook of public sector economics},
  pages={375--406},
  year={2017},
  publisher={Routledge}
}

@incollection{fu2013statistical,
  title={Statistical analysis of the causes of bridge collapse in China},
  author={Fu, Zhongqiu and Ji, Bohai and Cheng, Miao and Maeno, Hirofumi},
  booktitle={Forensic Engineering 2012: Gateway to a Safer Tomorrow},
  pages={75--83},
  year={2013}
}

@article{wardhana2003analysis,
  title={Analysis of recent bridge failures in the United States},
  author={Wardhana, Kumalasari and Hadipriono, Fabian C},
  journal={Journal of performance of constructed facilities},
  volume={17},
  number={3},
  pages={144--150},
  year={2003},
  publisher={American Society of Civil Engineers}
}

@article{zhang2022causes,
  title={Causes and statistical characteristics of bridge failures: A review},
  author={Zhang, Guojing and Liu, Yongjian and Liu, Jiang and Lan, Shiyong and Yang, Jian},
  journal={Journal of traffic and transportation engineering (English edition)},
  volume={9},
  number={3},
  pages={388--406},
  year={2022},
  publisher={Elsevier}
}

@misc{DW2024Dresden,
  author       = {{Deutsche Welle}},
  title        = {Germany: Bridge in Dresden Collapses into {Elbe} River},
  howpublished = {Deutsche Welle (DW)},
  year         = {2024},
  month        = sep,
  day          = {11},
  url          = {https://www.dw.com/en/germany-bridge-in-dresden-collapses-into-elbe-river/a-70185172},
  note         = {Accessed: 2026}
}

@article{domaneschi2020collapse,
  title={Collapse analysis of the Polcevera viaduct by the applied element method},
  author={Domaneschi, Marco and Pellecchia, C and De Iuliis, E and Cimellaro, GP and Morgese, M and Khalil, AA and Ansari, F},
  journal={Engineering Structures},
  volume={214},
  pages={110659},
  year={2020},
  publisher={Elsevier}
}

@article{national2008collapse,
  title={Collapse of i-35w highway bridge, minneapolis, minnesota, august 1, 2007},
  author={National Transportation Safety Board},
  journal={Accident Report NTSB/HAR-08/03, PB2008916203},
  year={2008}
}

@inproceedings{aksaray2016q,
  title={Q-learning for robust satisfaction of signal temporal logic specifications},
  author={Aksaray, Derya and Jones, Austin and Kong, Zhaodan and Schwager, Mac and Belta, Calin},
  booktitle={2016 IEEE 55th Conference on Decision and Control (CDC)},
  pages={6565--6570},
  year={2016},
  organization={IEEE}
}

@inproceedings{li2017reinforcement,
  title={Reinforcement learning with temporal logic rewards},
  author={Li, Xiao and Vasile, Cristian-Ioan and Belta, Calin},
  booktitle={2017 IEEE/RSJ International Conference on Intelligent Robots and Systems (IROS)},
  pages={3834--3839},
  year={2017},
  organization={IEEE}
}

@article{le2024reinforcement,
  title={Reinforcement learning with LTL and omega-regular objectives via optimality-preserving translation to average rewards},
  author={Le, Xuan Bach and Wagner, Dominik and Witzman, Leon and Rabinovich, Alexander and Ong, Luke},
  journal={Advances in Neural Information Processing Systems},
  volume={37},
  pages={117109--117132},
  year={2024}
}

@inproceedings{alur2023policy,
  title={Policy synthesis and reinforcement learning for discounted ltl},
  author={Alur, Rajeev and Bastani, Osbert and Jothimurugan, Kishor and Perez, Mateo and Somenzi, Fabio and Trivedi, Ashutosh},
  booktitle={International Conference on Computer Aided Verification},
  pages={415--435},
  year={2023},
  organization={Springer}
}

@inproceedings{cai2021reinforcement,
  title={Reinforcement learning based temporal logic control with maximum probabilistic satisfaction},
  author={Cai, Mingyu and Xiao, Shaoping and Li, Baoluo and Li, Zhiliang and Kan, Zhen},
  booktitle={2021 IEEE International Conference on Robotics and Automation (ICRA)},
  pages={806--812},
  year={2021},
  organization={IEEE}
}

@article{hahn2020reward,
  title={Reward shaping for reinforcement learning with omega-regular objectives},
  author={Hahn, Ernst M and Perez, Mateo and Schewe, Sven and Somenzi, Fabio and Trivedi, Ashutosh and Wojtczak, Dominik},
  journal={arXiv preprint arXiv:2001.05977},
  year={2020}
}

@inproceedings{bozkurt2020control,
  title={Control synthesis from linear temporal logic specifications using model-free reinforcement learning},
  author={Bozkurt, Alper Kamil and Wang, Yu and Zavlanos, Michael M and Pajic, Miroslav},
  booktitle={2020 IEEE International Conference on Robotics and Automation (ICRA)},
  pages={10349--10355},
  year={2020},
  organization={IEEE}
}

@inproceedings{hasanbeig2019reinforcement,
  title={Reinforcement learning for temporal logic control synthesis with probabilistic satisfaction guarantees},
  author={Hasanbeig, Mohammadhosein and Kantaros, Yiannis and Abate, Alessandro and Kroening, Daniel and Pappas, George J and Lee, Insup},
  booktitle={2019 IEEE 58th conference on decision and control (CDC)},
  pages={5338--5343},
  year={2019},
  organization={IEEE}
}

@article{gross2025turn,
  title={Turn-based Multi-Agent Reinforcement Learning Model Checking},
  author={Gross, Dennis},
  journal={arXiv preprint arXiv:2501.03187},
  year={2025}
}

@article{shao2023sample,
  title={Sample efficient model-free reinforcement learning from LTL specifications with optimality guarantees},
  author={Shao, Daqian and Kwiatkowska, Marta},
  journal={arXiv preprint arXiv:2305.01381},
  year={2023}
}

@article{jansen2018machine,
  title={Machine learning and model checking join forces (dagstuhl seminar 18121)},
  author={Jansen, Nils and Katoen, Joost-Pieter and Kohli, Pusmeet and Kretinsky, Jan},
  journal={Dagstuhl Reports},
  volume={8},
  number={3},
  pages={74--93},
  year={2018},
  publisher={Schloss Dagstuhl--Leibniz-Zentrum f{\"u}r Informatik}
}

@article{landers2023deep,
  title={Deep reinforcement learning verification: A survey},
  author={Landers, Matthew and Doryab, Afsaneh},
  journal={ACM Computing Surveys},
  volume={55},
  number={14s},
  pages={1--31},
  year={2023},
  publisher={ACM New York, NY}
}

@article{schmittlearning,
  title={Learning Reactive Synthesis from Model Checking Feedback},
  author={Schmitt, Frederik and Cosler, Matthias and Ghanem, Mohamed and Krsmanovi{\'c}, Vladimir and Finkbeiner, Bernd}
}

@article{bensalem2024bridging,
  title={Bridging formal methods and machine learning with model checking and global optimisation},
  author={Bensalem, Saddek and Huang, Xiaowei and Ruan, Wenjie and Tang, Qiyi and Wu, Changshun and Zhao, Xingyu},
  journal={Journal of Logical and Algebraic Methods in Programming},
  volume={137},
  pages={100941},
  year={2024},
  publisher={Elsevier}
}

@article{thompson1998pontis,
  title={The Pontis bridge management system},
  author={Thompson, Paul D and Small, Edgar P and Johnson, Michael and Marshall, Allen R},
  journal={Structural engineering international},
  volume={8},
  number={4},
  pages={303--308},
  year={1998},
  publisher={Taylor \& Francis}
}

@unknown{MirzaeiZ,
author = {Mirzaei, Zanyar},
year = {2014},
month = {06},
pages = {},
title = {Overview of existing Bridge Management Systems - Report by the IABMAS Bridge Management Committee}
}

@article{vouros2022explainable,
  title={Explainable deep reinforcement learning: state of the art and challenges},
  author={Vouros, George A},
  journal={ACM Computing Surveys},
  volume={55},
  number={5},
  pages={1--39},
  year={2022},
  publisher={ACM New York, NY}
}

@article{hawk1995bridgit,
  title={BRIDGIT deterioration models},
  author={Hawk, Hugh},
  journal={Transportation Research Record},
  volume={1490},
  pages={19},
  year={1995},
  publisher={NATIONAL ACADEMY OF SCIENCES}
}

@article{DBLP:journals/corr/MnihKSGAWR13,
  author       = {Volodymyr Mnih and
                  Koray Kavukcuoglu and
                  David Silver and
                  Alex Graves and
                  Ioannis Antonoglou and
                  Daan Wierstra and
                  Martin A. Riedmiller},
  title        = {Playing Atari with Deep Reinforcement Learning},
  journal      = {CoRR},
  volume       = {abs/1312.5602},
  year         = {2013}
}

@article{andriotis2019managing,
  title={Managing engineering systems with large state and action spaces through deep reinforcement learning},
  author={Andriotis, Charalampos P and Papakonstantinou, Konstantinos G},
  journal={Reliability Engineering \& System Safety},
  volume={191},
  pages={106483},
  year={2019},
  publisher={Elsevier}
}

@misc{gross2026coolmcverifyingexplainingrl,
      title={COOL-MC: Verifying and Explaining RL Policies for Platelet Inventory Management}, 
      author={Dennis Gross},
      year={2026},
      eprint={2603.02396},
      archivePrefix={arXiv},
      primaryClass={cs.AI},
      url={https://arxiv.org/abs/2603.02396}, 
}

@inproceedings{hu2025reward,
  title={Reward Hacking in Reinforcement Learning and RLHF: A Multidisciplinary Examination of Vulnerabilities, Mitigation Strategies, and Alignment Challenges},
  author={Hu, Tiechuan and Zhu, Wenbo and Yan, Yuqi},
  booktitle={2025 5th Intelligent Cybersecurity Conference (ICSC)},
  pages={272--275},
  year={2025},
  organization={IEEE}
}

@misc{gross2026formallyverifyingexplainingsepsis,
      title={Formally Verifying and Explaining Sepsis Treatment Policies with COOL-MC}, 
      author={Dennis Gross},
      year={2026},
      eprint={2602.14505},
      archivePrefix={arXiv},
      primaryClass={cs.AI},
      url={https://arxiv.org/abs/2602.14505}, 
}

@inproceedings{gross2025pctl,
  title={PCTL Model Checking for Temporal RL Policy Safety Explanations},
  author={Gross, Dennis and Spieker, Helge},
  booktitle={Proceedings of the 40th ACM/SIGAPP Symposium on Applied Computing},
  pages={1514--1521},
  year={2025}
}

@inproceedings{DBLP:conf/aips/GrossS0023,
  author       = {Dennis Gross and
                  Christoph Schmidl and
                  Nils Jansen and
                  Guillermo A. P{\'{e}}rez},
  title        = {Model Checking for Adversarial Multi-Agent Reinforcement Learning
                  with Reactive Defense Methods},
  booktitle    = {{ICAPS}},
  pages        = {162--170},
  publisher    = {{AAAI} Press},
  year         = {2023}
}

@inproceedings{DBLP:conf/icaart/GrossS0023,
  author       = {Dennis Gross and
                  Thiago D. Sim{\~{a}}o and
                  Nils Jansen and
                  Guillermo A. P{\'{e}}rez},
  title        = {Targeted Adversarial Attacks on Deep Reinforcement Learning Policies
                  via Model Checking},
  booktitle    = {{ICAART} {(3)}},
  pages        = {501--508},
  publisher    = {{SCITEPRESS}},
  year         = {2023}
}

@article{graybeal2002visual,
  title={Visual inspection of highway bridges},
  author={Graybeal, Benjamin A and Phares, Brent M and Rolander, Dennis D and Moore, Mark and Washer, Glenn},
  journal={Journal of nondestructive evaluation},
  volume={21},
  number={3},
  pages={67--83},
  year={2002},
  publisher={Springer}
}

@article{nair2010acoustic,
  title={Acoustic emission monitoring of bridges: Review and case studies},
  author={Nair, Archana and Cai, CS},
  journal={Engineering structures},
  volume={32},
  number={6},
  pages={1704--1714},
  year={2010},
  publisher={Elsevier}
}

@article{zhang2022towards,
  title={Towards fully automated unmanned aerial vehicle-enabled bridge inspection: Where are we at?},
  author={Zhang, Cheng and Zou, Yang and Wang, Feng and del Rey Castillo, Enrique and Dimyadi, Johannes and Chen, Long},
  journal={Construction and Building Materials},
  volume={347},
  pages={128543},
  year={2022},
  publisher={Elsevier}
}

@article{wei2019reinforcement,
  title={Reinforcement learning in the maintenance of civil infrastructures},
  author={Wei, Shiyin and Jin, Xiaowei and Li, Hui},
  year={2019}
}

@article{yang2020life,
  title={Life-cycle management of deteriorating bridge networks with network-level risk bounds and system reliability analysis},
  author={Yang, David Y and Frangopol, Dan M},
  journal={Structural Safety},
  volume={83},
  pages={101911},
  year={2020},
  publisher={Elsevier}
}

@article{gopal1991application,
  title={Application of Markov decision process to level-of-service-based maintenance systems},
  author={Gopal, Siva and Majidzadeh, Kamran},
  journal={Transportation Research Record},
  number={1304},
  year={1991}
}

@article{morato2022optimal,
  title={Optimal inspection and maintenance planning for deteriorating structural components through dynamic Bayesian networks and Markov decision processes},
  author={Morato, Pablo G and Papakonstantinou, Konstantinos G and Andriotis, Charalampos P and Nielsen, Jannie S{\o}nderk{\ae}r and Rigo, Philippe},
  journal={Structural Safety},
  volume={94},
  pages={102140},
  year={2022},
  publisher={Elsevier}
}

@article{papakonstantinou2014optimum,
  title={Optimum inspection and maintenance policies for corroded structures using partially observable Markov decision processes and stochastic, physically based models},
  author={Papakonstantinou, KG and Shinozuka, M},
  journal={Probabilistic Engineering Mechanics},
  volume={37},
  pages={93--108},
  year={2014},
  publisher={Elsevier}
}

@article{robelin2007history,
  title={History-dependent bridge deck maintenance and replacement optimization with Markov decision processes},
  author={Robelin, Charles-Antoine and Madanat, Samer M},
  journal={Journal of infrastructure systems},
  volume={13},
  number={3},
  pages={195--201},
  year={2007},
  publisher={American Society of Civil Engineers}
}

@article{ellis1995inspection,
  title={Inspection, maintenance, and repair with partial observability},
  author={Ellis, Hugh and Jiang, Mingxiang and Corotis, Ross B},
  journal={Journal of Infrastructure Systems},
  volume={1},
  number={2},
  pages={92--99},
  year={1995},
  publisher={American Society of Civil Engineers}
}

@article{tao1994reliability,
  title={Reliability-based bridge design and life cycle management with Markov decision processes},
  author={Tao, Zongwei and Corotis, Ross B and Ellis, J Hugh},
  journal={Structural Safety},
  volume={16},
  number={1-2},
  pages={111--132},
  year={1994},
  publisher={Elsevier}
}

@article{jimenez2024maintenance,
  title={Maintenance strategies for sewer pipes with multi-state degradation and deep reinforcement learning},
  author={Jimenez-Roa, Lisandro A and Sim{\~a}o, Thiago D and Bukhsh, Zaharah and Tinga, Tiedo and Molegraaf, Hajo and Jansen, Nils and Stoelinga, Mari{\"e}lle},
  journal={arXiv preprint arXiv:2407.12894},
  year={2024}
}

@article{madanat1994optimal,
  title={Optimal inspection and repair policies for infrastructure facilities},
  author={Madanat, Samer and Ben-Akiva, Moshe},
  journal={Transportation science},
  volume={28},
  number={1},
  pages={55--62},
  year={1994},
  publisher={INFORMS}
}

@article{scherer1994markovian,
  title={Markovian models for bridge maintenance management},
  author={Scherer, William T and Glagola, Douglas M},
  journal={Journal of transportation engineering},
  volume={120},
  number={1},
  pages={37--51},
  year={1994},
  publisher={American Society of Civil Engineers}
}

@article{lee2024stochastic,
  title={A stochastic track maintenance scheduling model based on deep reinforcement learning approaches},
  author={Lee, Jun S and Yeo, In-Ho and Bae, Younghoon},
  journal={Reliability Engineering \& System Safety},
  volume={241},
  pages={109709},
  year={2024},
  publisher={Elsevier}
}

@inproceedings{botteghi2021towards,
  title={Towards autonomous pipeline inspection with hierarchical reinforcement learning},
  author={Botteghi, Nicolo and Grefte, Luuk and Poel, Mannes and Sirmacek, Beril and Brune, Christoph and Dertien, Edwin and Stramigioli, Stefano},
  booktitle={International Conference on Robot Intelligence Technology and Applications},
  pages={259--271},
  year={2021},
  organization={Springer}
}

@article{arcieri2024pomdp,
  title={POMDP inference and robust solution via deep reinforcement learning: An application to railway optimal maintenance},
  author={Arcieri, Giacomo and Hoelzl, Cyprien and Schwery, Oliver and Straub, Daniel and Papakonstantinou, Konstantinos G and Chatzi, Eleni},
  journal={Machine Learning},
  volume={113},
  number={10},
  pages={7967--7995},
  year={2024},
  publisher={Springer}
}

@article{sresakoolchai2023railway,
  title={Railway infrastructure maintenance efficiency improvement using deep reinforcement learning integrated with digital twin based on track geometry and component defects},
  author={Sresakoolchai, Jessada and Kaewunruen, Sakdirat},
  journal={Scientific reports},
  volume={13},
  number={1},
  pages={2439},
  year={2023},
  publisher={Nature Publishing Group UK London}
}

@article{mohammadi2022deep,
  title={A deep reinforcement learning approach for rail renewal and maintenance planning},
  author={Mohammadi, Reza and He, Qing},
  journal={Reliability Engineering \& System Safety},
  volume={225},
  pages={108615},
  year={2022},
  publisher={Elsevier}
}

@article{li2024research,
  title={Research on pavement maintenance planning using reinforcement learning},
  author={Li, Hui and Wang, Shuo and Pan, Jie and Shan, Fei and Jiang, Yuzhe and Zhang, Haopeng and Dai, Zhenhua and Shi, Qingjiang},
  journal={Intelligent Transportation Infrastructure},
  volume={3},
  pages={liaf001},
  year={2024},
  publisher={Oxford University Press}
}

@article{kazemeini2023identifying,
  title={Identifying environmentally sustainable pavement management strategies via deep reinforcement learning},
  author={Kazemeini, Ali and Swei, Omar},
  journal={Journal of cleaner production},
  volume={390},
  pages={136124},
  year={2023},
  publisher={Elsevier}
}

@article{yao2022large,
  title={Large-scale maintenance and rehabilitation optimization for multi-lane highway asphalt pavement: a reinforcement learning approach},
  author={Yao, Linyi and Leng, Zhen and Jiang, Jiwang and Ni, Fujian},
  journal={IEEE Transactions on Intelligent Transportation Systems},
  volume={23},
  number={11},
  pages={22094--22105},
  year={2022},
  publisher={IEEE}
}

@article{barua2022planning,
  title={Planning maintenance and rehabilitation activities for airport pavements: A combined supervised machine learning and reinforcement learning approach},
  author={Barua, Limon and Zou, Bo},
  journal={International Journal of Transportation Science and Technology},
  volume={11},
  number={2},
  pages={423--435},
  year={2022},
  publisher={Elsevier}
}

@article{latifi2021deep,
  title={A deep reinforcement learning model for predictive maintenance planning of road assets: Integrating LCA and LCCA},
  author={Latifi, Moein and Darvishvand, Fateme Golivand and Khandel, Omid and Nowsoud, Mobin Latifi},
  journal={arXiv preprint arXiv:2112.12589},
  year={2021}
}

@article{renard2021minimizing,
  title={Minimizing the global warming impact of pavement infrastructure through reinforcement learning},
  author={Renard, Sophie and Corbett, Benjamin and Swei, Omar},
  journal={Resources, conservation and recycling},
  volume={167},
  pages={105240},
  year={2021},
  publisher={Elsevier}
}

@article{han2021asphalt,
  title={Asphalt pavement maintenance plans intelligent decision model based on reinforcement learning algorithm},
  author={Han, Chengjia and Ma, Tao and Chen, Siyu},
  journal={Construction and Building Materials},
  volume={299},
  pages={124278},
  year={2021},
  publisher={Elsevier}
}

@article{yao2020deep,
  title={Deep reinforcement learning for long-term pavement maintenance planning},
  author={Yao, Linyi and Dong, Qiao and Jiang, Jiwang and Ni, Fujian},
  journal={Computer-Aided Civil and Infrastructure Engineering},
  volume={35},
  number={11},
  pages={1230--1245},
  year={2020},
  publisher={Wiley Online Library}
}

@article{asghari2025network,
  title={Network-Level Infrastructure Asset Management with Multiagent Actor--Critic Reinforcement Learning: A Case of Highway Bridges},
  author={Asghari, Vahid and Jahanbiglari, Ava and Hsu, Shu-Chien},
  journal={Journal of Infrastructure Systems},
  volume={31},
  number={4},
  pages={04025026},
  year={2025},
  publisher={American Society of Civil Engineers}
}

@inproceedings{liu2025multi,
  title={Multi-environment reinforcement learning for decision-making on bridge maintenance with varying interval},
  author={Liu, Gang and Jiang, Liming and Yang, Qingshan and Law, SS and Sun, Ruiqing and Zhao, Chen},
  booktitle={Structures},
  volume={80},
  pages={109778},
  year={2025},
  organization={Elsevier}
}

@article{thangeda2024infralib,
  title={InfraLib: Enabling Reinforcement Learning and Decision-Making for Large-Scale Infrastructure Management},
  author={Thangeda, Pranay and Betz, Trevor S and Grussing, Michael N and Ornik, Melkior},
  journal={arXiv preprint arXiv:2409.03167},
  year={2024}
}

@article{jing2026intelligent,
  title={Intelligent e-Maintenance for long-span bridges with preference-based decision making using A3C reinforcement learning},
  author={Jing, Qiang and Zhang, Jiaxin and Lai, Li and Dong, You},
  journal={Engineering Structures},
  volume={346},
  pages={121636},
  year={2026},
  publisher={Elsevier}
}

@article{van2025deep,
  title={Deep multi-objective reinforcement learning for utility-based infrastructural maintenance optimization},
  author={van Remmerden, Jesse and Kenter, Maurice and Roijers, Diederik M and Andriotis, Charalampos and Zhang, Yingqian and Bukhsh, Zaharah},
  journal={Neural Computing and Applications},
  volume={37},
  number={30},
  pages={24719--24742},
  year={2025},
  publisher={Springer}
}

@article{cheng2024knowledge,
  title={Knowledge transfer for adaptive maintenance policy optimization in engineering fleets based on meta-reinforcement learning},
  author={Cheng, Jianda and Cheng, Minghui and Liu, Yan and Wu, Jun and Li, Wei and Frangopol, Dan M},
  journal={Reliability Engineering \& System Safety},
  volume={247},
  pages={110127},
  year={2024},
  publisher={Elsevier}
}

@article{saifullah2024multi,
  title={Multi-agent deep reinforcement learning with centralized training and decentralized execution for transportation infrastructure management},
  author={Saifullah, M and Papakonstantinou, KG and Andriotis, CP and Stoffels, SM},
  journal={arXiv preprint arXiv:2401.12455},
  year={2024}
}

@article{lai2024synergetic,
  title={Synergetic-informed deep reinforcement learning for sustainable management of transportation networks with large action spaces},
  author={Lai, Li and Dong, You and Andriotis, Charalampos P and Wang, Aijun and Lei, Xiaoming},
  journal={Automation in Construction},
  volume={160},
  pages={105302},
  year={2024},
  publisher={Elsevier}
}

@article{taherkhani2024towards,
  title={Towards equitable infrastructure asset management: Scour maintenance strategy for aging bridge systems in flood-prone zones using deep reinforcement learning},
  author={Taherkhani, Amir and Mo, Weiwei and Bell, Erin and Han, Fei},
  journal={Sustainable Cities and Society},
  volume={114},
  pages={105792},
  year={2024},
  publisher={Elsevier}
}

@article{hamida2023hierarchical,
  title={Hierarchical reinforcement learning for transportation infrastructure maintenance planning},
  author={Hamida, Zachary and Goulet, James-A},
  journal={Reliability Engineering \& System Safety},
  volume={235},
  pages={109214},
  year={2023},
  publisher={Elsevier}
}

@article{morato2023inference,
  title={Inference and dynamic decision-making for deteriorating systems with probabilistic dependencies through Bayesian networks and deep reinforcement learning},
  author={Morato, Pablo G and Andriotis, Charalampos P and Papakonstantinou, Konstantinos G and Rigo, Philippe},
  journal={Reliability Engineering \& System Safety},
  volume={235},
  pages={109144},
  year={2023},
  publisher={Elsevier}
}

@article{lei2023sustainable,
  title={Sustainable life-cycle maintenance policymaking for network-level deteriorating bridges with a convolutional autoencoder--structured reinforcement learning agent},
  author={Lei, Xiaoming and Dong, You and Frangopol, Dan M},
  journal={Journal of Bridge Engineering},
  volume={28},
  number={9},
  pages={04023063},
  year={2023},
  publisher={American Society of Civil Engineers}
}

@article{dong2022deep,
  title={Deep reinforcement learning for optimal life-cycle management of deteriorating regional bridges using double-deep Q-networks},
  author={Dong, You and Lei, Xiaoming},
  journal={Smart Structures and Systems},
  pages={571--582},
  year={2022}
}

@article{zhou2022reinforcement,
  title={A reinforcement learning method for multiasset roadway improvement scheduling considering traffic impacts},
  author={Zhou, Weiwen and Miller-Hooks, Elise and Papakonstantinou, Kostas G and Stoffels, Shelley and McNeil, Sue},
  journal={Journal of Infrastructure Systems},
  volume={28},
  number={4},
  pages={04022033},
  year={2022},
  publisher={American Society of Civil Engineers}
}

@article{zhou2022advanced,
  title={An advanced multi-agent reinforcement learning framework of bridge maintenance policy formulation},
  author={Zhou, Qi-Neng and Yuan, Ye and Yang, Dong and Zhang, Jing},
  journal={Sustainability},
  volume={14},
  number={16},
  pages={10050},
  year={2022},
  publisher={MDPI}
}

@article{du2022parameterized,
  title={Parameterized deep reinforcement learning-enabled maintenance decision-support and life-cycle risk assessment for highway bridge portfolios},
  author={Du, Ao and Ghavidel, Alireza},
  journal={Structural Safety},
  volume={97},
  pages={102221},
  year={2022},
  publisher={Elsevier}
}

@article{lei2022deep,
  title={A deep reinforcement learning framework for life-cycle maintenance planning of regional deteriorating bridges using inspection data},
  author={Lei, Xiaoming and Xia, Ye and Deng, Lu and Sun, Limin},
  journal={Structural and Multidisciplinary Optimization},
  volume={65},
  number={5},
  pages={149},
  year={2022},
  publisher={Springer}
}

@article{yang2022deep,
  title={Deep reinforcement learning--enabled bridge management considering asset and network risks},
  author={Yang, David Y},
  journal={Journal of Infrastructure Systems},
  volume={28},
  number={3},
  pages={04022023},
  year={2022},
  publisher={American Society of Civil Engineers}
}

@article{yang2022adaptive,
  title={Adaptive risk-based life-cycle management for large-scale structures using deep reinforcement learning and surrogate modeling},
  author={Yang, David Y},
  journal={Journal of Engineering Mechanics},
  volume={148},
  number={1},
  pages={04021126},
  year={2022},
  publisher={American Society of Civil Engineers}
}

@article{cheng2021decision,
  title={A decision-making framework for load rating planning of aging bridges using deep reinforcement learning},
  author={Cheng, Minghui and Frangopol, Dan M},
  journal={Journal of Computing in Civil Engineering},
  volume={35},
  number={6},
  pages={04021024},
  year={2021},
  publisher={American Society of Civil Engineers}
}

@incollection{robelin2006dynamic,
  title={Dynamic programming based maintenance and replacement optimization for bridge decks using history-dependent deterioration models},
  author={Robelin, Charles-Antoine and Madanat, Samer M},
  booktitle={Applications of advanced technology in transportation},
  pages={13--18},
  year={2006}
}

@article{medury2014simultaneous,
  title={Simultaneous network optimization approach for pavement management systems},
  author={Medury, Aditya and Madanat, Samer},
  journal={Journal of Infrastructure Systems},
  volume={20},
  number={3},
  pages={04014010},
  year={2014},
  publisher={American Society of Civil Engineers}
}

@article{kuhn2010network,
  title={Network-level infrastructure management using approximate dynamic programming},
  author={Kuhn, Kenneth D},
  journal={Journal of Infrastructure Systems},
  volume={16},
  number={2},
  pages={103--111},
  year={2010},
  publisher={American Society of Civil Engineers}
}

@article{ahmed2020review,
  title={Review of non-destructive civil infrastructure evaluation for bridges: State-of-the-art robotic platforms, sensors and algorithms},
  author={Ahmed, Habib and La, Hung Manh and Gucunski, Nenad},
  journal={Sensors},
  volume={20},
  number={14},
  pages={3954},
  year={2020},
  publisher={MDPI}
}

@article{matos2023comparison,
  title={Comparison of condition rating systems for bridges in three European countries},
  author={Matos, Jos{\'e} C and Nicoletti, Vanni and Kralovanec, Jakub and Sousa, H{\'e}lder S and Gara, Fabrizio and Moravcik, Martin and Morais, Maria J},
  journal={Applied Sciences},
  volume={13},
  number={22},
  pages={12343},
  year={2023},
  publisher={MDPI}
}

@techreport{fhwa2022nbis,
  author      = {{Federal Highway Administration}},
  title       = {National Bridge Inspection Standards: Final Rule},
  institution = {U.S. Department of Transportation, Federal Highway Administration},
  year        = {2022},
  number      = {FHWA-2017-0047},
  note        = {87 FR 27396, 23 CFR Part 650, effective June 6, 2022},
  address     = {Washington, D.C.},
  url         = {https://www.federalregister.gov/documents/2022/05/06/2022-09512/national-bridge-inspection-standards}
}

@techreport{fhwa1995recording,
  author      = {{Federal Highway Administration}},
  title       = {Recording and Coding Guide for the Structure Inventory 
                 and Appraisal of the Nation's Bridges},
  institution = {U.S. Department of Transportation, Federal Highway Administration},
  year        = {1995},
  number      = {FHWA-PD-96-001},
  address     = {Washington, D.C.},
  url         = {https://www.fhwa.dot.gov/bridge/mtguide.pdf}
}

@article{DBLP:journals/ml/Bekkemoen24,
  author       = {Yanzhe Bekkemoen},
  title        = {Explainable reinforcement learning {(XRL):} a systematic literature
                  review and taxonomy},
  journal      = {Mach. Learn.},
  volume       = {113},
  number       = {1},
  pages        = {355--441},
  year         = {2024}
}

@article{samadi2024safe,
  title={SAFE-RL: Saliency-aware counterfactual explainer for deep reinforcement learning policies},
  author={Samadi, Amir and Koufos, Konstantinos and Debattista, Kurt and Dianati, Mehrdad},
  journal={IEEE Robotics and Automation Letters},
  volume={9},
  number={11},
  pages={9994--10001},
  year={2024},
  publisher={IEEE}
}

@inproceedings{DBLP:conf/aiide/SieusahaiG21,
  author       = {Alexander Sieusahai and
                  Matthew Guzdial},
  title        = {Explaining Deep Reinforcement Learning Agents in the Atari Domain
                  through a Surrogate Model},
  booktitle    = {{AIIDE}},
  pages        = {82--90},
  publisher    = {{AAAI} Press},
  year         = {2021}
}

@inproceedings{DBLP:conf/esann/GrossS24,
  author       = {Dennis Gross and
                  Helge Spieker},
  title        = {Safety-Oriented Pruning and Interpretation of Reinforcement Learning
                  Policies},
  booktitle    = {{ESANN}},
  year         = {2024}
}

@article{schulman2017proximal,
  title={Proximal policy optimization algorithms},
  author={Schulman, John and Wolski, Filip and Dhariwal, Prafulla and Radford, Alec and Klimov, Oleg},
  journal={arXiv preprint arXiv:1707.06347},
  year={2017}
}

@inproceedings{DBLP:conf/icaart/GrossS25,
  author       = {Dennis Gross and
                  Helge Spieker},
  title        = {Co-Activation Graph Analysis of Safety-Verified and Explainable Deep
                  Reinforcement Learning Policies},
  booktitle    = {{ICAART} {(2)}},
  pages        = {611--621},
  publisher    = {{SCITEPRESS}},
  year         = {2025}
}

@inproceedings{gross2024enhancing,
  title={Enhancing rl safety with counterfactual llm reasoning},
  author={Gross, Dennis and Spieker, Helge},
  booktitle={IFIP International Conference on Testing Software and Systems},
  pages={23--29},
  year={2024},
  organization={Springer}
}

@inproceedings{DBLP:conf/pldi/ZhuXMJ19,
  author       = {He Zhu and
                  Zikang Xiong and
                  Stephen Magill and
                  Suresh Jagannathan},
  title        = {An inductive synthesis framework for verifiable reinforcement learning},
  booktitle    = {{PLDI}},
  pages        = {686--701},
  publisher    = {{ACM}},
  year         = {2019}
}

@inproceedings{DBLP:conf/seke/JinWZ22,
  author       = {Peng Jin and
                  Yang Wang and
                  Min Zhang},
  title        = {Efficient {LTL} Model Checking of Deep Reinforcement Learning Systems
                  using Policy Extraction},
  booktitle    = {{SEKE}},
  pages        = {357--362},
  publisher    = {{KSI} Research Inc.},
  year         = {2022}
}

@inproceedings{DBLP:conf/sigcomm/EliyahuKKS21,
  author       = {Tomer Eliyahu and
                  Yafim Kazak and
                  Guy Katz and
                  Michael Schapira},
  title        = {Verifying learning-augmented systems},
  booktitle    = {{SIGCOMM}},
  pages        = {305--318},
  publisher    = {{ACM}},
  year         = {2021}
}

@inproceedings{DBLP:conf/sigcomm/KazakBKS19,
  author       = {Yafim Kazak and
                  Clark W. Barrett and
                  Guy Katz and
                  Michael Schapira},
  title        = {Verifying Deep-RL-Driven Systems},
  booktitle    = {NetAI@SIGCOMM},
  pages        = {83--89},
  publisher    = {{ACM}},
  year         = {2019}
}

@article{DBLP:journals/csur/MilaniTVF24,
  author       = {Stephanie Milani and
                  Nicholay Topin and
                  Manuela Veloso and
                  Fei Fang},
  title        = {Explainable Reinforcement Learning: {A} Survey and Comparative Review},
  journal      = {{ACM} Comput. Surv.},
  volume       = {56},
  number       = {7},
  pages        = {168:1--168:36},
  year         = {2024}
}

@article{littman2017environment,
  title={Environment-independent task specifications via GLTL},
  author={Littman, Michael L and Topcu, Ufuk and Fu, Jie and Isbell, Charles and Wen, Min and MacGlashan, James},
  journal={arXiv preprint arXiv:1704.04341},
  year={2017}
}

@inproceedings{sadigh2014learning,
  title={A learning based approach to control synthesis of markov decision processes for linear temporal logic specifications},
  author={Sadigh, Dorsa and Kim, Eric S and Coogan, Samuel and Sastry, S Shankar and Seshia, Sanjit A},
  booktitle={53rd IEEE Conference on Decision and Control},
  pages={1091--1096},
  year={2014},
  organization={IEEE}
}

@article{fu2014probably,
  title={Probably approximately correct MDP learning and control with temporal logic constraints},
  author={Fu, Jie and Topcu, Ufuk},
  journal={arXiv preprint arXiv:1404.7073},
  year={2014}
}

@inproceedings{wang2024safe,
  title={Safe exploration in reinforcement learning by reachability analysis over learned models},
  author={Wang, Yuning and Zhu, He},
  booktitle={International Conference on Computer Aided Verification},
  pages={232--255},
  year={2024},
  organization={Springer}
}

@inproceedings{hunt2021verifiably,
  title={Verifiably safe exploration for end-to-end reinforcement learning},
  author={Hunt, Nathan and Fulton, Nathan and Magliacane, Sara and Hoang, Trong Nghia and Das, Subhro and Solar-Lezama, Armando},
  booktitle={Proceedings of the 24th International Conference on Hybrid Systems: Computation and Control},
  pages={1--11},
  year={2021}
}

@article{ganai2023iterative,
  title={Iterative reachability estimation for safe reinforcement learning},
  author={Ganai, Milan and Gong, Zheng and Yu, Chenning and Herbert, Sylvia and Gao, Sicun},
  journal={Advances in Neural Information Processing Systems},
  volume={36},
  pages={69764--69797},
  year={2023}
}

@inproceedings{fisac2019bridging,
  title={Bridging hamilton-jacobi safety analysis and reinforcement learning},
  author={Fisac, Jaime F and Lugovoy, Neil F and Rubies-Royo, Vicen{\c{c}} and Ghosh, Shromona and Tomlin, Claire J},
  booktitle={2019 International Conference on Robotics and Automation (ICRA)},
  pages={8550--8556},
  year={2019},
  organization={IEEE}
}

@article{fisac2018general,
  title={A general safety framework for learning-based control in uncertain robotic systems},
  author={Fisac, Jaime F and Akametalu, Anayo K and Zeilinger, Melanie N and Kaynama, Shahab and Gillula, Jeremy and Tomlin, Claire J},
  journal={IEEE Transactions on Automatic Control},
  volume={64},
  number={7},
  pages={2737--2752},
  year={2018},
  publisher={IEEE}
}

@inproceedings{unniyankal2023rmlgym,
  title={RMLGym: a Formal Reward Machine Framework for Reinforcement Learning.},
  author={Unniyankal, Hisham and Belardinelli, Francesco and Ferrando, Angelo and Malvone, Vadim},
  booktitle={WOA},
  pages={1--16},
  year={2023}
}

@article{marzari2025verifying,
  title={Verifying Online Safety Properties for Safe Deep Reinforcement Learning},
  author={Marzari, Luca and Cicalese, Ferdinando and Farinelli, Alessandro and Amato, Christopher and Marchesini, Enrico},
  journal={ACM Transactions on Intelligent Systems and Technology},
  volume={17},
  number={1},
  pages={1--27},
  year={2025},
  publisher={ACM New York, NY}
}

@article{zolfagharian2024smarla,
  title={Smarla: A safety monitoring approach for deep reinforcement learning agents},
  author={Zolfagharian, Amirhossein and Abdellatif, Manel and Briand, Lionel C and others},
  journal={IEEE Transactions on Software Engineering},
  volume={51},
  number={1},
  pages={82--105},
  year={2024},
  publisher={IEEE}
}

@article{mannucci2023runtime,
  title={Runtime verification of learning properties for reinforcement learning algorithms},
  author={Mannucci, Tommaso and others},
  journal={arXiv preprint arXiv:2311.09811},
  year={2023}
}

@inproceedings{lazarus2020runtime,
  title={Runtime safety assurance using reinforcement learning},
  author={Lazarus, Christopher and Lopez, James G and Kochenderfer, Mykel J},
  booktitle={2020 AIAA/IEEE 39th Digital Avionics Systems Conference (DASC)},
  pages={1--9},
  year={2020},
  organization={IEEE}
}

@inproceedings{mallozzi2019runtime,
  title={A runtime monitoring framework to enforce invariants on reinforcement learning agents exploring complex environments},
  author={Mallozzi, Piergiuseppe and Castellano, Ezequiel and Pelliccione, Patrizio and Schneider, Gerardo and Tei, Kenji},
  booktitle={2019 IEEE/ACM 2nd International Workshop on Robotics Software Engineering (RoSE)},
  pages={5--12},
  year={2019},
  organization={IEEE}
}

@inproceedings{jin2022trainify,
  title={Trainify: A cegar-driven training and verification framework for safe deep reinforcement learning},
  author={Jin, Peng and Tian, Jiaxu and Zhi, Dapeng and Wen, Xuejun and Zhang, Min},
  booktitle={International Conference on Computer Aided Verification},
  pages={193--218},
  year={2022},
  organization={Springer}
}

@article{wu2024verified,
  title={Verified safe reinforcement learning for neural network dynamic models},
  author={Wu, Junlin and Zhang, Huan and Vorobeychik, Yevgeniy},
  journal={Advances in Neural Information Processing Systems},
  volume={37},
  pages={117762--117783},
  year={2024}
}

@article{gangopadhyay2021counterexample,
  title={Counterexample guided RL policy refinement using bayesian optimization},
  author={Gangopadhyay, Briti and Dasgupta, Pallab},
  journal={Advances in Neural Information Processing Systems},
  volume={34},
  pages={22783--22794},
  year={2021}
}

@article{akintunde2022formal,
  title={Formal verification of neural agents in non-deterministic environments},
  author={Akintunde, Michael E and Botoeva, Elena and Kouvaros, Panagiotis and Lomuscio, Alessio},
  journal={Autonomous Agents and Multi-Agent Systems},
  volume={36},
  number={1},
  pages={6},
  year={2022},
  publisher={Springer}
}

@inproceedings{bacci2020probabilistic,
  title={Probabilistic guarantees for safe deep reinforcement learning},
  author={Bacci, Edoardo and Parker, David},
  booktitle={International Conference on Formal Modeling and Analysis of Timed Systems},
  pages={231--248},
  year={2020},
  organization={Springer}
}

@inproceedings{dong2022dependability,
  title={Dependability analysis of deep reinforcement learning based robotics and autonomous systems through probabilistic model checking},
  author={Dong, Yi and Zhao, Xingyu and Huang, Xiaowei},
  booktitle={2022 IEEE/RSJ International Conference on Intelligent Robots and Systems (IROS)},
  pages={5171--5178},
  year={2022},
  organization={IEEE}
}

@inproceedings{bacci2022verified,
  title={Verified probabilistic policies for deep reinforcement learning},
  author={Bacci, Edoardo and Parker, David},
  booktitle={NASA Formal Methods Symposium},
  pages={193--212},
  year={2022},
  organization={Springer}
}

@inproceedings{bacci2021verifying,
  title={Verifying reinforcement learning up to infinity},
  author={Bacci, Edoardo and Giacobbe, Mirco and Parker, David},
  booktitle={Proceedings of the International Joint Conference on Artificial Intelligence},
  year={2021},
  organization={International Joint Conferences on Artificial Intelligence Organization}
}

@article{jin2021learning,
  title={Learning on abstract domains: A new approach for verifiable guarantee in reinforcement learning},
  author={Jin, Peng and Zhang, Min and Li, Jianwen and Han, Li and Wen, Xuejun},
  journal={arXiv preprint arXiv:2106.06931},
  year={2021}
}

@article{tian2023boosting,
  title={Boosting verification of deep reinforcement learning via piece-wise linear decision neural networks},
  author={Tian, Jiaxu and Zhi, Dapeng and Liu, Si and Wang, Peixin and Chen, Cheng and Zhang, Min},
  journal={Advances in Neural Information Processing Systems},
  volume={36},
  pages={10022--10037},
  year={2023}
}

@inproceedings{corsi2021formal,
  title={Formal verification of neural networks for safety-critical tasks in deep reinforcement learning},
  author={Corsi, Davide and Marchesini, Enrico and Farinelli, Alessandro},
  booktitle={Uncertainty in Artificial Intelligence},
  pages={333--343},
  year={2021},
  organization={PMLR}
}

@article{tran2019safety,
  title={Safety verification of cyber-physical systems with reinforcement learning control},
  author={Tran, Hoang-Dung and Cai, Feiyang and Diego, Manzanas Lopez and Musau, Patrick and Johnson, Taylor T and Koutsoukos, Xenofon},
  journal={ACM Transactions on Embedded Computing Systems (TECS)},
  volume={18},
  number={5s},
  pages={1--22},
  year={2019},
  publisher={ACM New York, NY, USA}
}

@inproceedings{venkataraman2020tractable,
  title={Tractable reinforcement learning of signal temporal logic objectives},
  author={Venkataraman, Harish and Aksaray, Derya and Seiler, Peter},
  booktitle={Learning for Dynamics and Control},
  pages={308--317},
  year={2020},
  organization={PMLR}
}

@article{anderson2020neurosymbolic,
  title={Neurosymbolic reinforcement learning with formally verified exploration},
  author={Anderson, Greg and Verma, Abhinav and Dillig, Isil and Chaudhuri, Swarat},
  journal={Advances in neural information processing systems},
  volume={33},
  pages={6172--6183},
  year={2020}
}

@inproceedings{junges2016safety,
  title={Safety-constrained reinforcement learning for MDPs},
  author={Junges, Sebastian and Jansen, Nils and Dehnert, Christian and Topcu, Ufuk and Katoen, Joost-Pieter},
  booktitle={International conference on tools and algorithms for the construction and analysis of systems},
  pages={130--146},
  year={2016},
  organization={Springer}
}

@book{baier2008principles,
  title     = {Principles of model checking},
  author    = {Baier, Christel and Katoen, Joost-Pieter},
  year      = {2008},
  publisher = {MIT press}
}

@misc{prism_manual, 
title="{PRISM Manual}",
url={https://www.prismmodelchecker.org/doc/semantics.pdf},
author = "{PRISM}",
year = "{2023}",
howpublished = "\url{www.prismmodelchecker.org}",
note = "Accessed: {03/14/2024}"
}

@article{DBLP:journals/corr/DragerFK0U15,
  author    = {Klaus Dr{\"{a}}ger and
               Vojtech Forejt and
               Marta Z. Kwiatkowska and
               David Parker and
               Mateusz Ujma},
  title     = {Permissive Controller Synthesis for Probabilistic Systems},
  journal   = {Log. Methods Comput. Sci.},
  volume    = {11},
  number    = {2},
  year      = {2015}
}

@article{hansson1994logic,
  title={A logic for reasoning about time and reliability},
  author={Hansson, Hans and Jonsson, Bengt},
  journal={Formal aspects of computing},
  volume={6},
  number={5},
  pages={512--535},
  year={1994},
  publisher={Springer}
}

@article{DBLP:journals/sttt/HenselJKQV22,
  author    = {Christian Hensel and
               Sebastian Junges and
               Joost{-}Pieter Katoen and
               Tim Quatmann and
               Matthias Volk},
  title     = {The probabilistic model checker {Storm}},
  journal   = {Int. J. Softw. Tools Technol. Transf.},
  volume    = {24},
  number    = {4},
  pages     = {589--610},
  year      = {2022}
}

@inproceedings{DBLP:conf/setta/GrossJJP22,
  author       = {Dennis Gross and
                  Nils Jansen and
                  Sebastian Junges and
                  Guillermo A. P{\'{e}}rez},
  title        = {{COOL-MC:} {A} Comprehensive Tool for Reinforcement Learning and Model
                  Checking},
  booktitle    = {{SETTA}},
  series       = {Lecture Notes in Computer Science},
  volume       = {13649},
  pages        = {41--49},
  publisher    = {Springer},
  year         = {2022}
}

@article{lu2024overcoming,
  title={Overcoming the curse of dimensionality in reinforcement learning through approximate factorization},
  author={Lu, Chenbei and Shi, Laixi and Chen, Zaiwei and Wu, Chenye and Wierman, Adam},
  journal={arXiv preprint arXiv:2411.07591},
  year={2024}
}
